\documentclass[lettersize,journal]{IEEEtran}
\usepackage{amsmath,amsfonts}
\usepackage{algorithmic}
\usepackage{algorithm}
\usepackage{array}
\usepackage[caption=false,font=normalsize,labelfont=sf,textfont=sf]{subfig}
\usepackage{textcomp}
\usepackage{stfloats}
\usepackage{url}
\usepackage{verbatim}
\usepackage{graphicx}
\usepackage{cite}
\usepackage{bm}
\usepackage{colortbl}
\usepackage{threeparttable}
\usepackage{booktabs}
\usepackage{amssymb}
\usepackage{xcolor}
\usepackage{multirow}

\makeatletter
\let\MYcaption\@makecaption
\makeatother

\usepackage{subcaption}
\makeatletter
\let\@makecaption\MYcaption
\makeatother

\newcommand{\MYheader}{\smash{\scriptsize
\hfil\parbox[t][\height][t]{\textwidth}{\centering
This paper has been published in IEEE Transactions on Instrumentation and Measurement.
Digital Object Identifier (DOI) is \protect\url{https://doi.org/10.1109/TIM.2026.3687304}}\hfil\hbox{}}}

\newcommand{\MYfooter}{\smash{\scriptsize
\hfil\parbox[t][\height][t]{\textwidth}{\centering
© 2026 IEEE.  Personal use of this material is permitted.  Permission from IEEE must be obtained for all other uses, in any current or future media, including reprinting/republishing this material for advertising or promotional purposes, creating new collective works, for resale or redistribution to servers or lists, or reuse of any copyrighted component of this work in other works.}\hfil\hbox{}}}

\makeatletter

\def\ps@headings{%
\def\@oddhead{\mbox{}\scriptsize\MYheader \hfil \thepage}
\def\@evenhead{\scriptsize\thepage \hfil \MYheader\mbox{}}
\def\@oddfoot{\MYfooter}%
\def\@evenfoot{\MYfooter}}

\def\ps@IEEEtitlepagestyle{%
\def\@oddhead{\mbox{}\scriptsize\MYheader \hfil \thepage}%
\def\@evenhead{\scriptsize\thepage \hfil \MYheader\mbox{}}%
\def\@oddfoot{\MYfooter}%
\def\@evenfoot{\MYfooter}}

\makeatother

\begin{document}

\title{U-FaceBP: Uncertainty-aware Bayesian Ensemble Deep Learning\\ for Face Video-based Blood Pressure Estimation}

\author{Yusuke Akamatsu, Akinori F. Ebihara, and Terumi Umematsu,~\IEEEmembership{Senior Member,~IEEE,}
        % <-this % stops a space
\thanks{Yusuke Akamatsu, Akinori F. Ebihara, and Terumi Umematsu are with Biometrics Research Laboratories, NEC Corporation, Kawasaki, Japan (e-mail: yusuke-akamatsu@nec.com, aebihara@nec.com, terumi@nec.com).}% <-this % stops a space
%\thanks{Manuscript received April 19, 2021; revised August 16, 2021.}}
}

% \markboth{IEEE Transactions on Instrumentation and Measurement,~Vol.~XX, No.~XX, XX~2026}%


\maketitle

\begin{abstract}
Blood pressure (BP) measurement is crucial for daily health assessment.
Remote photoplethysmography (rPPG), which extracts pulse waves from face videos captured by a camera, has the potential to enable convenient BP measurement without specialized medical devices.
However, there are various uncertainties in BP estimation using rPPG, leading to limited estimation performance and reliability.
In this paper, we propose U-FaceBP, an uncertainty-aware Bayesian ensemble deep learning method for face video-based BP estimation.
U-FaceBP models aleatoric and epistemic uncertainties in face video-based BP estimation with a Bayesian neural network (BNN). 
Additionally, we design U-FaceBP as an ensemble method, estimating BP from rPPG signals, PPG signals derived from face videos, and face images using multiple BNNs.
Large-scale experiments on two datasets involving 1197 subjects from diverse racial groups demonstrate that U-FaceBP outperforms state-of-the-art BP estimation methods.
Furthermore, we show that the uncertainty estimates provided by U-FaceBP are informative and useful for guiding modality fusion, assessing prediction reliability, and analyzing performance across racial groups.
\end{abstract}

\begin{IEEEkeywords}
Blood pressure, Remote photoplethysmography, Uncertainty-aware deep learning, Bayesian neural network, Facial analysis.
\end{IEEEkeywords}

\section{Introduction}
\IEEEPARstart{B}{lood} pressure (BP) is a fundamental vital sign for assessing human health. 
According to the World Health Organization (WHO)~\cite{world2022world}, the number of adults with hypertension ($i.e.$, high BP) has reached 1.28 billion. 
Hypertension significantly increases the risk of heart, brain, and kidney disease.
Notably, 30--50\% of hypertensive individuals are unaware of their condition~\cite{lugo2017factors}, highlighting the importance of regular BP monitoring.
Traditional BP measurement relies on mercury or electronic sphygmomanometers, which use a cuff to compress the arm and detect the pressure exerted by blood.
However, the discomfort and inconvenience of cuff-based measurement discourage many individuals---especially those who take little care of their health---from regularly monitoring their BP. 

\begin{figure}[t]
\begin{center}
\includegraphics[scale=0.39]{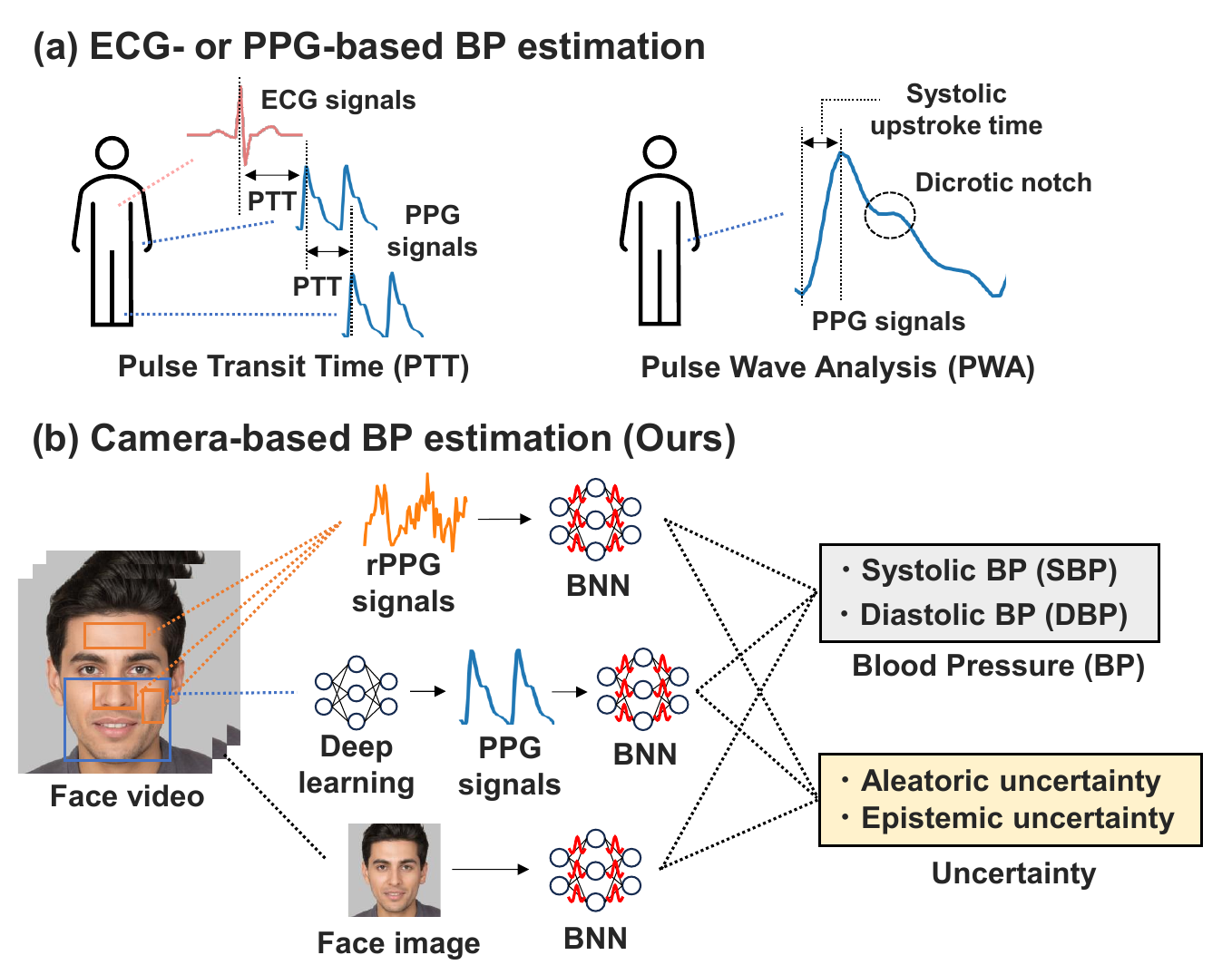}
\end{center}
\vspace{-10pt}
\caption{(a) PTT- and PWA-based methods for cuffless BP estimation using ECG or PPG. (b) U-FaceBP for camera-based BP estimation. We design U-FaceBP as an ensemble method consisting of rPPG, PPG signals (reconstructed from face videos), and face image-based BP estimation models. U-FaceBP outputs predicted BP and its uncertainty.}
\label{fig:intro}
\end{figure}

To address this, \textit{cuffless BP estimation methods} based on electrocardiography (ECG) and photoplethysmography (PPG) have been explored.
These methods can be categorized into two main approaches (see Fig.~\ref{fig:intro}~(a)): pulse transit time (PTT)-based~\cite{chan2001noninvasive,gesche2012continuous,kachuee2016cuffless,yang2020estimation,block2020conventional} and pulse wave analysis (PWA)-based~\cite{teng2003continuous,kurylyak2013neural,xing2016optical,slapnivcar2019blood,leitner2021personalized} methods. 
PTT-based methods leverage the inverse correlation between PTT and BP~\cite{geddes1981pulse}, requiring ECG and PPG or multiple PPG sensors to estimate PTT. 
However, these methods necessitate multiple sensors attached to different body parts ($e.g.$, chest and fingertips, toes and fingertips~\cite{block2020conventional}), making them impractical for casual use. 
PWA-based methods estimate BP using a single PPG sensor by analyzing waveform characteristics ($e.g.$, systolic upstroke time, dicrotic notch). 
Despite their simplicity, PPG sensors are not widely available in home environments and require skin contact, which is not always readily available outside of clinical settings.

Recently, camera-based BP estimation from face videos has emerged as a promising alternative to PPG-based approaches~\cite{frey2022blood,man2022blood}. 
This method enables \textit{non-contact BP monitoring} using commonly available webcams and smartphones, making it particularly useful for individuals who do not measure their BP regularly.
BP estimation from face videos primarily relies on remote photoplethysmography (rPPG) signals~\cite{luo2019smartphone,zhou2019noninvasive,rong2021blood,schrumpf2021assessment,wu2022facial,li2022hybrid,wu2022camera,liu2023vidbp,trirongjitmoah2024assessing,cheng2025remote}, which are extracted from subtle color variations in the skin. 
An alternative approach derives PTT features from multiple rPPG signals captured from different facial regions, $e.g.$, cheeks and forehead~\cite{wu2022facial,wu2023contactless,park2024robust}. 
Additionally, some methods~\cite{wu2022facial,wu2023contactless} incorporate age and body mass index (BMI) estimated from face images, as these factors are correlated with BP~\cite{droyvold2005change}.

Despite its advantages, BP estimation from face videos presents several challenges. 
First, rPPG signals are inherently weak and highly susceptible to noise from body motion and lighting variations, degrading estimation performance~\cite{trirongjitmoah2024assessing}. 
Second, while pulse wave features are correlated with BP, the validity of BP estimation from pulse waves remains low~\cite{mehta2024examining}. 
Furthermore, the cardiovascular system is highly individualized, and the performance of BP estimation from rPPG signals depends on the person~\cite{schrumpf2021assessment}.
These factors introduce various \textit{uncertainties}, leading to limited estimation performance and reliability.

To address these challenges, we propose \textbf{U-FaceBP}~\footnote{This paper reuses some content from thesis~\cite{yuaka2025thesis} with permission.}, an uncertainty-aware Bayesian ensemble deep learning method for face video-based BP estimation (see Fig.~\ref{fig:intro}~(b)). 
U-FaceBP models aleatoric and epistemic uncertainties using a Bayesian neural network (BNN), which explicitly accounts for uncertainty in deep learning models. 
Furthermore, U-FaceBP employs an ensemble approach that integrates multiple BP-related features:

\begin{itemize}
\item \textbf{rPPG-based BP estimation}: Multiple rPPG signals extracted from different facial regions are used to estimate BP using both pulse waveform and PTT features.
\item \textbf{PPG-based BP estimation}: Instead of relying on traditional rPPG signal processing techniques ($e.g.$, Plane Orthogonal-to-Skin (POS)~\cite{wang2016algorithmic}), we use deep learning to estimate \textit{contact PPG signals} from face videos, which improves the signal quality and extracts the precise pulse waveform.
\item \textbf{Face image-based BP estimation}: Facial appearance features ($e.g.$, age~\cite{o2007mechanical}, gender~\cite{oparil2005gender}, BMI~\cite{droyvold2005change}, facial shape~\cite{stephen2017facial,ang2021novel}, wrinkles, and baldness~\cite{lin2020feasibility,liu2022videocad} related to BP) extracted from face images are used for BP estimation.
\end{itemize}
The predictions from the above three perspectives are integrated into a final predicted BP via an \textit{uncertainty-driven aggregator (UDA)}.
To enhance reliability and transparency, U-FaceBP explicitly models two types of uncertainty:

\begin{itemize}
\item \textbf{Aleatoric uncertainty}: Noise inherent in the observations, such as uncertainty of BP estimation from pulse waves and individual physiological differences.
\item \textbf{Epistemic uncertainty}: Lack of knowledge of the model, $i.e.$, the model exhibits high uncertainty for unfamiliar data not included in the training data.
\end{itemize}
By leveraging these uncertainties, U-FaceBP provides prediction confidence, enabling a triage-like approach: reliable or clearly healthy cases can be handled by AI, while uncertain or potentially abnormal cases are assessed by human experts or a conventional BP monitor. 
This division of labor improves both efficiency and safety in practical deployment.

We conduct large-scale experiments on two datasets with 1197 subjects from diverse racial groups, ensuring practical evaluation settings by preventing subject overlap between training and test data. 
Our results demonstrate that U-FaceBP surpasses state-of-the-art (SoTA) BP estimation methods in both estimation performance and reliability.

The main contributions of this paper are as follows:

\begin{itemize}
\item \textbf{Uncertainty-aware framework}: We leverage uncertainty for three key purposes: (i) \textit{Modality fusion}, by guiding modality integration through UDA; (ii) \textit{Confidence assessment}, by identifying reliable predictions; and (iii) \textit{Racial subgroup analysis}, by revealing performance disparities across racial groups.
\item \textbf{Novel ensemble approach}: We integrate BP estimation from rPPG, PPG signals (reconstructed from face videos), and face images, achieving superior performance compared to prior methods relying solely on rPPG signals.
\item \textbf{Large-scale experimental validation}: We evaluate U-FaceBP on two datasets with 1197 racially diverse subjects, assessing its generalizability by ensuring a wide BP range and subject-independent testing.
\end{itemize}

\begin{table*}[tbp]
\centering
\caption{Overview comparison of blood pressure measurement technologies. Methods are categorized into seven types, including invasive, cuff-based, wearable, cuffless, and contactless. Our method belongs to camera-based contactless methods, as indicated in gray. $\bigstar$, $\bigcirc$, and $\triangle$ indicate established, high/moderate, and limited levels, respectively.}
\vspace{-5pt}
\label{table:bp-tech}
\small
\resizebox{\textwidth}{!}{%
\begin{tabular}{p{2.8cm} p{2.5cm} p{2.5cm} p{2cm} p{2.5cm} p{2cm} p{2cm} p{2.2cm} p{2.5cm}}
\noalign{\hrule height 1.3pt} 
\hline
\noalign{\smallskip}
\centering \textbf{Category} & 
\centering \textbf{Measurement Principle} & 
\centering \textbf{Ease of Use} & 
\centering \textbf{Regulatory Approval} & 
\centering \textbf{Accuracy} & 
\centering \textbf{Installation Burden} & 
\centering \textbf{Price Range} & 
\centering \textbf{Market Stage} & 
{\centering \textbf{Key Features}\par} \\
\noalign{\smallskip}
\hline
\noalign{\smallskip}

\centering \textbf{Invasive Arterial Line} &
\centering Direct intra-arterial catheter measurement &
\centering $\triangle$ \\ Requires invasive procedure &
\centering $\bigstar$ \\ ICU standard &
\centering $\bigstar$ \\ Gold standard &
\centering $\triangle$ \\ Very high &
\centering $\triangle$ \\ High overall cost &
\centering Clinical standard (ICU / Operating Room) &
{\centering Highest accuracy; critical care\par} \\
\noalign{\smallskip}
\hline
\noalign{\smallskip}

\centering \textbf{Non-invasive Cuff (Upper Arm)} &
\centering Oscillometric method &
\centering $\bigcirc$ \\ Cuff attachment required &
\centering $\bigstar$ \\ Approved &
\centering $\bigcirc$ \\ Highly accurate (device dependent) &
\centering $\bigcirc$ \\ Low burden &
\centering $\bigcirc$ \\ Low cost &
\centering Widely used in home \& clinical settings &
{\centering Clinical standard\par} \\
\noalign{\smallskip}
\hline
\noalign{\smallskip}

\centering \textbf{Continuous Non-invasive (e.g., CNAP~\footnotemark[2], Finapres~\footnotemark[3])} &
\centering Volume clamp method &
\centering $\triangle$ \\ Some restriction/discomfort &
\centering $\bigstar$ \\ Approved &
\centering $\bigcirc$ \\ Highly accurate (device dependent) &
\centering $\triangle$ \\ High &
\centering $\triangle$ \\ Expensive &
\centering Implemented in ICU \& Operating room &
{\centering Continuous intraoperative monitoring\par} \\
\noalign{\smallskip}
\hline
\noalign{\smallskip}

\centering \textbf{Wearable Blood Pressure Monitor (e.g., HeartGuide~\footnotemark[4])} &
\centering Mini cuff &
\centering $\bigcirc$ \\ Suitable for daily wear &
\centering $\bigcirc$ \\ Device dependent approval &
\centering $\bigcirc$ \\ Accurate (device dependent) &
\centering $\bigcirc$ \\ Low burden &
\centering $\bigcirc$ \\ Moderate cost &
\centering Expanding consumer market &
{\centering Daily monitoring\par} \\
\noalign{\smallskip}
\hline
\noalign{\smallskip}

\centering \textbf{Cuffless (PPG + ECG)} &
\centering Pulse Transit Time (PTT) analysis &
\centering $\bigcirc$ \\ Lightweight sensors &
\centering $\triangle$ \\ Limited approval &
\centering $\bigcirc$ \\ Moderately accurate &
\centering $\bigcirc$ \\ Moderate burden &
\centering $\bigcirc$ \\ Moderate cost &
\centering Early commercialization stage &
{\centering Continuous estimation oriented\par} \\
\noalign{\smallskip}
\hline
\noalign{\smallskip}

\centering \textbf{Cuffless (PPG only)} &
\centering Pulse waveform + Machine learning &
\centering $\bigstar$ \\ Very simple &
\centering $\triangle$ \\ Limited approval &
\centering $\triangle$ \\ Needs improvement &
\centering $\bigcirc$ \\ Low burden &
\centering $\bigcirc$ \\ Low cost &
\centering Research / pilot stage &
{\centering Simple device configuration\par} \\
\noalign{\smallskip}
\hline
\noalign{\smallskip}

\rowcolor[rgb]{0.9, 0.9, 0.9}
\centering \textbf{Contactless (Camera-based)} &
\centering rPPG-based estimation &
\centering $\bigstar$ \\ Completely contactless &
\centering $\triangle$ \\ Limited approval &
\centering $\triangle$ \\ Needs improvement &
\centering $\bigstar$ \\ Easy setup &
\centering $\bigstar$ \\ No dedicated hardware &
\centering Research / pilot stage &
{\centering Remote measurement \& Scalable \par} \\
\noalign{\smallskip}
\hline
\noalign{\hrule height 1.3pt}

\end{tabular}
}
\end{table*}

\section{Related Work}
\label{sec:related_work}

\subsection{Blood Pressure Measurement Technologies}
Various technologies have been developed for BP measurement, including invasive, cuff-based, wearable, cuffless, and contactless approaches.
TABLE~\ref{table:bp-tech} compares seven types of BP measurement technologies, summarizing their principles, ease of use, approvals, accuracy, burden and cost, market stage, and features.
Invasive approaches are clinically reliable, but they involve high overall costs and impose a significant burden on patients.
The cuff-based approach is non-invasive and highly accurate, making it the most widely used method for BP measurement currently.
Continuous non-invasive systems provide beat-to-beat monitoring and may serve as a less invasive alternative to arterial lines in selected clinical scenarios, although device cost remains high.
The wearable BP monitor is equipped with a small cuff on the device, making it suitable for daily BP measurements.
The cuffless approach using PPG and ECG offers moderate accuracy but requires two sensors and tends to involve complex configurations.
The cuffless approach using only PPG has a very simple configuration but requires further accuracy improvement.
The camera-based contactless approach is low-burden without physical contact and eliminates the need for dedicated hardware by utilizing smartphones, significantly expanding opportunities for BP measurement.
Such methods have the potential to facilitate remote measurements such as telemedicine and simultaneous measurements of multiple individuals.
Nevertheless, they remain in the research stage, and further validation and accuracy improvements are necessary.
The proposed method falls within this camera-based contactless category and aims to enhance estimation performance and reliability.

\subsection{Face Video-based Blood Pressure Estimation}
Face video-based physiological measurement has advanced significantly over the past 15 years~\cite{poh2010non,de2013robust,wang2016algorithmic}, enabling non-contact estimation of vital signs such as heart rate and respiratory rate.  
rPPG extracts pulse waves from subtle changes in reflected light on the skin and has recently been explored for BP estimation~\cite{frey2022blood}.  
Luo et al.~\cite{luo2019smartphone} pioneered large-scale BP estimation from face videos, extracting rPPG signals via transdermal optical imaging and then estimating BP using a multilayer perceptron.  
However, their method required additional meta-features ($e.g.$, room temperature, age, skin tone).  
Schrumpf et al.~\cite{schrumpf2021assessment} trained a convolutional neural network (CNN) to estimate BP from PPG signals then conducted transfer learning to estimate BP from rPPG signals.
They applied a personalization technique, which re-trains the CNN with person-specific data, to improve BP estimation performance.
Although previous methods often require physical characteristics ($e.g.$, age, BMI)~\cite{luo2019smartphone,zhou2019noninvasive,li2022hybrid} or personalization with person-specific data~\cite{schrumpf2021assessment,wu2022camera,liu2023vidbp}, U-FaceBP does not use them.
While Han et al.~\cite{han2023noncontact} are the only ones to provide uncertainty in BP estimation from rPPG signals, their approach is limited to post-hoc confidence assessment for a single modality.
In contrast, we are the first to exploit uncertainty as a unified framework for modality fusion, confidence assessment, and racial subgroup analysis in BP estimation from face videos.
Additionally, U-FaceBP integrates multiple BP estimation strategies, including deep learning-based PPG reconstruction and direct face image-based BP prediction.  

\subsection{Uncertainty Estimation}  
Uncertainty estimation~\cite{gawlikowski2023survey} is essential for improving transparency and reliability of a neural network (NN), particularly in high-risk fields like medical imaging~\cite{nair2020exploring} and autonomous driving~\cite{feng2018towards}.  
A common way to estimate uncertainty is BNNs~\cite{blundell2015weight,gal2015bayesian,kendall2017uncertainties,krueger2017bayesian,mobiny2021dropconnect}. 
BNNs are classified into three different types based on how the posterior distribution is inferred to approximate Bayesian inference~\cite{gawlikowski2023survey}: variational inference~\cite{hinton1993keeping} such as Monte Carlo (MC) dropout~\cite{gal2016dropout}, sampling approaches such as Markov Chain Monte Carlo sampling (MCMC)~\cite{neal1992bayesian}, and Laplace approximation~\cite{denker1990transforming,daxberger2021laplace}.
Kendall et al.~\cite{kendall2017uncertainties} introduced BNN using MC dropout to quantify aleatoric uncertainty and epistemic uncertainty together in a single model.  
Another way is an ensemble approach~\cite{lakshminarayanan2017simple,valdenegro2019deep,wenbatchensemble}.
Lakshminarayanan et al.~\cite{lakshminarayanan2017simple} showed that model diversity, induced by random initialization and shuffled training data, enables effective uncertainty estimation.  

\footnotetext[2]{\url{https://www.cnsystems.com/technology/cnap-blood-pressure/}}
\footnotetext[3]{\url{https://www.finapres.com/}}
\footnotetext[4]{\url{https://www.omron-healthcare.com/products/heartguide}}

\section{Methodology}
\begin{figure*}[t]
\begin{center}
\includegraphics[scale=0.29]{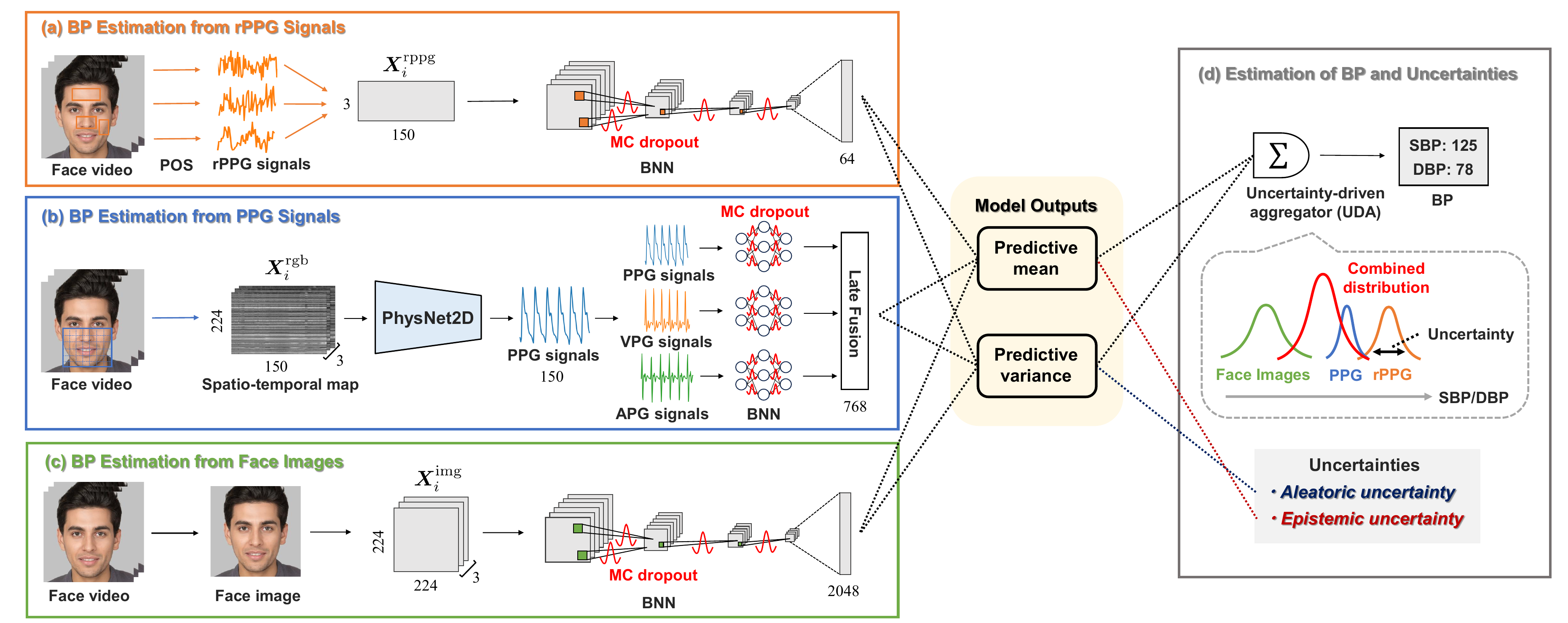}
\end{center}
\vspace{-13pt}
\caption{Framework of U-FaceBP: (a) BP estimation from rPPG signals, (b) BP estimation from PPG signals, (c) BP estimation from face images, and (d) estimation of uncertainties and BP. Face photo by Generated Photos (https://generated.photos/).}
\vspace{-5pt}
\label{fig:method}
\end{figure*}

The framework of U-FaceBP is illustrated in Fig.~\ref{fig:method}.
We first introduce uncertainty in Bayesian deep learning.
We then explain BP estimation from rPPG signals, PPG signals (reconstructed from face videos), and face images.
Finally, we explain the estimation of uncertainties and BP.

\subsection{Uncertainty in Bayesian Deep Learning}
\label{subsec:uncertainty}
U-FaceBP introduces aleatoric and epistemic uncertainties using BNNs~\cite{kendall2017uncertainties}.
To capture epistemic uncertainty in an NN, a BNN sets a prior distribution over its weights, such as a Gaussian prior distribution: ${\bf W}\sim \mathcal{N}(0,I)$.
This replaces the deterministic weight parameters in the NN with distributions over those parameters.
Denoting the random output of the BNN as $f^{{\bf W}}({\bf x})$ and ground-truth BP as $y$, the model likelihood is defined as $p(y|f^{{\bf W}}({\bf x}))$.
Since BP estimation is a regression task, U-FaceBP defines the likelihood as a Gaussian distribution: $p(y|f^{{\bf W}}({\bf x}))=\mathcal{N}(\mu({\bf x}),\sigma({\bf x})^2)$.
Here, $\mu({\bf x})$ is the predictive mean when ${\bf x}$ is input into the model.
Also, $\sigma({\bf x})^2$ is the observation noise (predictive variance) that represents aleatoric uncertainty depending on the input ${\bf x}$, referred to as heteroscedastic uncertainty.

Given a dataset with input ${\bf X}$ and output ${\bf Y}$, Bayesian inference is used to compute the posterior over the weights $p({\bf W|X,Y})=p({\bf Y|X,W})p({\bf W})/p({\bf Y|X})$.
Since $p({\bf Y|X})$ cannot be calculated analytically, the posterior $p({\bf W|X,Y})$ is approximated with a simple distribution $q_{\theta}^*({\bf W})$.
We use variational inference with MC dropout~\cite{gal2016dropout} to find $q_{\theta}^*({\bf W})$ that minimizes the Kullback-Leibler (KL) divergence to the true posterior $p({\bf W|X,Y})$.
MC dropout is conducted by training the model with dropout in each weight layer and by also conducting dropout during testing to sample from the approximate posterior.
U-FaceBP, as also an ensemble method, estimates BP from face videos with multiple BNNs.
In the following sections, we describe how this BNN is adapted to each ensemble network in our framework.

\subsection{BP Estimation from rPPG Signals}
\label{subsec:rppg}

As a training dataset, we use a set of face videos ${\bf X}$, systolic BP (SBP) ${\bf Y_{\rm sbp}}$, and diastolic BP (DBP) ${\bf Y_{\rm dbp}}$, denoted as $({\bf X},{\bf Y_{\rm sbp}},{\bf Y_{\rm dbp}})=\{{\bf X}_i, y_{{\rm sbp},i},y_{{\rm dbp},i}\}_{i=1}^{N}$, where $i$ is the sample index and $N$ is the number of samples.
We extract rPPG signals from face videos then estimate BP on the basis of the BNN.
We first apply face detection~\cite{imaoka2021future} to the face videos and detect face feature points.
Then, rPPG signals are extracted from multiple regions of interests (ROIs) on the face.
Following~\cite{wu2022facial}, we extract rPPG signal $\bm{X}_i^{\rm rppg} \in \mathbb{R}^{K \times F}$ from three ROIs on the cheek, inner cheek, and forehead (see Fig.~\ref{fig:method}~(a)) using POS~\cite{wang2016algorithmic}.
Note that $K$ is the number of ROIs and $F$ is the number of frames ($K=3$ and $F=150$ in our experiment).
By extracting rPPG signals in multiple ROIs, we can use both pulse waveform and PTT features.
Specifically, the pulse waveform of rPPG signals contains BP-related features~\cite{wu2022camera}, and rPPG signals from multiple ROIs ($e.g.$, cheek and forehead) contain PTT-related features~\cite{wu2022facial} (see Appendix~A for actual data).

BP is then estimated from rPPG signals $\bm{X}_i^{\rm rppg}$ using the BNN.
To model both epistemic and aleatoric uncertainty, we extend S-Net~\cite{wu2022facial} with dropout layers after each residual block and configure it to output predictive means and variances for both SBP and DBP.
Specifically, the network outputs four values: SBP's predictive mean $\hat{y}_{{\rm sbp},i}^{\rm rppg}$ and variance $(\hat{\sigma}_{{\rm sbp},i}^{\rm rppg})^2$ and DBP's predictive mean $\hat{y}_{{\rm dbp},i}^{\rm rppg}$ and variance $(\hat{\sigma}_{{\rm dbp},i}^{\rm rppg})^2$.
We apply MC dropout at inference time to sample from the approximate posterior distribution.
The model is trained using the negative log-likelihood (NLL) loss~\cite{kendall2017uncertainties}:
\begin{eqnarray}
    \mathcal{L}_{\rm rppg}  =  \frac{1}{N} \sum_{i=1}^{N} \frac{||y_{{\rm sbp},i}-\hat{y}_{{\rm sbp},i}^{\rm rppg}||^2}{2(\hat{\sigma}_{{\rm sbp},i}^{\rm rppg})^2} +\frac{1}{2}\log(\hat{\sigma}_{{\rm sbp},i}^{\rm rppg})^2 \nonumber \\ + 
    \frac{1}{N} \sum_{i=1}^{N} \frac{||y_{{\rm dbp},i}-\hat{y}_{{\rm dbp},i}^{\rm rppg}||^2}{2(\hat{\sigma}_{{\rm dbp},i}^{\rm rppg})^2} +\frac{1}{2}\log(\hat{\sigma}_{{\rm dbp},i}^{\rm rppg})^2.
\end{eqnarray}
In the first terms for SBP (first row) and DBP (second row), the predictive variances $(\hat{\sigma}_{{\rm sbp},i}^{\rm rppg})^2$ and $(\hat{\sigma}_{{\rm dbp},i}^{\rm rppg})^2$ enable the model to be trained while handling aleatoric uncertainty.
More specifically, large predictive variances work to ignore regression loss, while small predictive variances work to emphasize regression loss.
The second terms $\frac{1}{2}\log(\hat{\sigma}_{{\rm sbp},i}^{\rm rppg})^2$ and $\frac{1}{2}\log(\hat{\sigma}_{{\rm dbp},i}^{\rm rppg})^2$ prevent the network from predicting infinite uncertainty ($i.e.$, zero loss) for all samples.
In practice, we train the network to predict the log variances $s_{{\rm sbp},i}^{\rm rppg}:=\log(\hat{\sigma}_{{\rm sbp},i}^{\rm rppg})^2$ and $s_{{\rm dbp},i}^{\rm rppg}:=\log(\hat{\sigma}_{{\rm dbp},i}^{\rm rppg})^2$ for numerical stability~\cite{kendall2017uncertainties}.
Unlike the original S-Net, our model explicitly incorporates both aleatoric and epistemic uncertainty, and serves as a key component within our uncertainty-aware ensemble framework.

\subsection{BP Estimation from PPG Signals}
\label{subsec:ppg}

The rPPG signals used in the previous section are hand-crafted features extracted through a signal processing approach~\cite{wang2016algorithmic}, which may result in lower-quality pulse waveforms.  
In contrast, deep learning-based PPG signal reconstruction from face videos has demonstrated robustness against noise caused by body motion and lighting variations~\cite{chen2018deepphys,liu2020multi,yu2022physformer}. 
To achieve precise and robust pulse waveform extraction, we reconstruct the PPG signal ${\bm p}_i \in \mathbb{R}^{D_{p}}$ from face videos using deep learning and subsequently estimate BP from the reconstructed PPG signal.  
Here, $D_{p}$ is the dimension of the PPG signal acquired from a contact-type PPG sensor ($D_{p}=150$ in our experiment).
It is important to note that rPPG signals contain PTT-related features, unlike PPG signals.  
Thus, rPPG and PPG signals provide complementary information for BP estimation.  

We first divide the facial region below the eyes into 16$\times$14 blocks (see Fig.~\ref{fig:method}~(b)) and construct a spatio-temporal map $\bm{X}_i^{\rm rgb} \in \mathbb{R}^{3 \times K \times F}$ using RGB time-series signals (3 being RGB channels, $K=224$, and $F=150$ in our experiment).
We use PhysNet2D~\cite{akamatsu2024calibrationphys} to estimate the PPG signal $\hat{\bm{p}}_i \in \mathbb{R}^{D_{p}}$ from the spatio-temporal map $\bm{X}_i^{\rm rgb}$.
As the loss function between $\hat{\bm{p}}_i$ and $\bm{p}_i$, we use the mean squared error (MSE) loss to minimize the error in the pulse waveform, which differs from the objective used in prior work~\cite{akamatsu2024calibrationphys}.
To further enhance physiological relevance, we use the first derivative (VPG signal) $\bm{p}^{\prime}_{i} \in \mathbb{R}^{D_{p}-1}$ and second derivative (APG signal) $\bm{p}^{\prime\prime}_{i} \in \mathbb{R}^{D_{p}-2}$ of the PPG signal, inspired by prior work on contact-based PPG analysis~\cite{liu2017cuffless,slapnivcar2019blood}.
These derivatives contain the information about aortic compliance and stiffness, which are closely associated with BP.
Thus, the PhysNet2D model is trained with the following multi-term loss:
$\mathcal{L}_{\rm pulse}  =  \frac{\alpha}{N} \sum_{i=1}^{N} ||\bm{p}_i - \hat{\bm{p}}_i||^2 + \frac{\beta}{N} \sum_{i=1}^{N} ||\bm{p}^{\prime}_i - \hat{\bm{p}}^{\prime}_i||^2
    + \frac{\gamma}{N} \sum_{i=1}^{N} ||\bm{p}^{\prime\prime}_i - \hat{\bm{p}}^{\prime\prime}_i||^2,$
where $\alpha$, $\beta$, and $\gamma$ are hyperparameters that control the weight of each loss, and $\hat{\bm{p}}^{\prime}_i$, $\hat{\bm{p}}^{\prime\prime}_i$ are computed from the estimated PPG signal.

BP is then estimated from the PPG $\hat{\bm{p}}_i$, VPG $\hat{\bm{p}}^{\prime}_i$, and APG signals $\hat{\bm{p}}^{\prime\prime}_i$ using the BNN.
We extend ResNet1D~\cite{schrumpf2021assessment}, which is used for BP estimation from PPG signals, into BNN by incorporating aleatoric and epistemic uncertainty.
U-FaceBP concatenates the feature representations obtained from each ResNet1D corresponding to PPG, VPG, and APG signals then outputs the predictive mean and variance (see Fig.~\ref{fig:method}~(b)).
The BNN is trained using the same NLL loss $\mathcal{L}_{\rm ppg}$ as in Eq.~(1).
Overall, U-FaceBP trains the network in an end-to-end fashion using the loss function $\mathcal{L}_{\rm pulse}+\mathcal{L}_{\rm ppg}$.
Unlike prior studies that rely solely on handcrafted rPPG signals, this is the first method to unify the estimation of PPG, VPG, and APG signals from face videos and BP prediction in an end-to-end manner, leveraging their physiological structure under uncertainty.

\subsection{BP Estimation from Face Images}
\label{subsec:image}

The face images contain BP-related information, such as age, BMI, facial shape, wrinkles, and baldness.
Wrinkles and baldness have been reported as signs associated with cardiovascular risk factors closely linked to hypertension~\cite{lin2020feasibility,liu2022videocad}.
The pulse waves and face images have complementary BP-related features, so the ensemble approach improves the BP estimation performance.
We obtain a face image $\bm{X}_i^{\rm img} \in \mathbb{R}^{3 \times H \times W}$ ($H$ and $W$ being the height and width of the image, respectively).
%Note that the images include the entire face to capture baldness.
We estimate BP from face images $\bm{X}_i^{\rm img}$ using a Bayesian extension of ResNet50~\cite{he2016deep} tailored for uncertainty-aware prediction.
Facial appearance features contain many uncertain factors such as individual differences, which motivates us to introduce this uncertainty into our model.
Since the number of face images paired with BPs for training data is limited, pre-training the BNN is essential.
We leverage a pre-trained model on the basis of facial representation learning (FRL)~\cite{bulat2022pre}, which is useful for various face tasks.
Specifically, FRL~\cite{bulat2022pre} performs SwAV~\cite{caron2020unsupervised}-based self-supervised learning using unlabeled face images to acquire feature representations related to human faces.
We then carry out fine-tuning ($i.e.$, transfer learning of all layers) of the BNN using the same NLL loss $\mathcal{L}_{\rm img}$ as in Eq.~(1).
This is the first attempt to enable direct and uncertainty-aware BP prediction from face images by fine-tuning a self-supervised facial representation model.

\subsection{Estimation of Uncertainties and BP}
\label{subsec:estimation}

We quantify uncertainty and estimate BP using an ensemble of multiple BNNs.
We explain this only for SBP, but the same applies to DBP.

By understanding the prediction confidence through uncertainty, we can flexibly handle the predicted BP.
We can also improve the transparency of deep learning models.
Suppose that we sample $T$ times according to MC dropout at test time. 
The aleatoric uncertainty for estimating the SBP from the rPPG signal is calculated as $U^{\rm aleatoric}_{\rm rppg, sbp} = \frac{1}{T} \sum_{t=1}^{T} (\hat{\sigma}_{{\rm sbp},t}^{\rm rppg})^2$, which is defined by the mean over $T$ samples of predictive variance used in Eq.~(1).
The epistemic uncertainty for estimating the SBP from the rPPG signal is calculated as
$U^{\rm epistemic}_{\rm rppg, sbp} = \frac{1}{T} \sum_{t=1}^{T} (\hat{y}_{{\rm sbp},t}^{\rm rppg})^2 - (\frac{1}{T} \sum_{t=1}^{T} \hat{y}_{{\rm sbp},t}^{\rm rppg} )^2$, which is defined as the variance of the predictive mean over $T$ samples.
These are the same for calculating the aleatoric and epistemic uncertainties $U^{\rm aleatoric}_{\rm ppg, sbp}$, $U^{\rm epistemic}_{\rm ppg, sbp}$ and $U^{\rm aleatoric}_{\rm img, sbp}$, $U^{\rm epistemic}_{\rm img, sbp}$ estimated from the PPG signal and face image, respectively.
We then calculate the total uncertainty for SBP as the sum of the uncertainties
$U^{\rm total}_{\rm sbp}  =  \sum_{m \in {{\{\rm rppg, ppg, img}}\}} U^{\rm aleatoric}_{m,\rm{sbp}} + U^{\rm epistemic}_{m,\rm{sbp}}$.
We can consider the predicted SBP to be highly reliable if the uncertainty $U^{\rm total}_{\rm sbp}$ is small, or vice versa.

Finally, we describe the prediction of SBP.
The predicted SBP from the rPPG signal is calculated as $\hat{y}_{{\rm sbp}}^{\rm rppg} = \frac{1}{T} \sum_{t=1}^{T} \hat{y}_{{\rm sbp},t}^{\rm rppg}$, where $\hat{y}_{{\rm sbp},t}^{\rm rppg}$ is the predictive mean at the $t$-th MC dropout sample.
Similarly, we compute $\hat{y}_{{\rm sbp}}^{\rm ppg}$ and $\hat{y}_{{\rm sbp}}^{\rm img}$ from the PPG signal and face image, respectively.
To aggregate these predictions, we propose UDA, which assigns higher weights to modalities with lower uncertainty.
UDA computes the final predicted SBP as $\hat{y}_{{\rm sbp}} = \sum_{m \in {{\{\rm rppg, ppg, img}}\}} {\rm softmax}(-s^{\rm total}_{m,\rm sbp})\cdot\hat{y}_{{\rm sbp}}^{m}$, where ${\rm softmax}(-s^{\rm total}_{m,\rm sbp})=\frac{{\rm exp}(-s^{\rm total}_{m,\rm sbp})}{\sum_{m' \in {{\{\rm rppg, ppg, img}}\}} {\rm exp}(-s^{\rm total}_{m',\rm sbp})}$ and $s^{\rm total}_{m,\rm sbp} = \log\left(\frac{U^{\rm aleatoric}_{m,\rm sbp}+U^{\rm epistemic}_{m,\rm sbp}}{\mu^{\rm aleatoric}_{m,\rm sbp} + \mu^{\rm epistemic}_{m,\rm sbp}}\right)$ is the log total uncertainty for $m$.
Each uncertainty term is normalized by the mean uncertainty $\mu$ for that modality, computed from the validation set, to correct for modality-wise scale differences.
This is the first framework that performs uncertainty-guided fusion of multiple face video-derived modalities for BP prediction.

\section{Experiments}
We conduct extensive experiments using two datasets, focusing on the pitfalls~\cite{mehta2024examining} of the experimental setting in BP estimation.
Mehta et al.~\cite{mehta2024examining} demonstrated that previous studies often misleadingly provide lower estimation errors.
Specifically, data leakage such as including the same subjects in both the training and test data, and task constraints such as limiting to subjects with normal BP ranges, have made the prediction task artificially ``easier".
Thus, we design our experiment on the basis of these points.

\setlength{\tabcolsep}{4pt}
\begin{table}[h]
\caption{Dataset statistics.}
\vspace{-10pt}
\label{table:stats}
\begin{center}
\scalebox{1}{
\begin{tabular}{lcc}
\noalign{\hrule height 1.3pt} \noalign{\smallskip}
 & FaceBP Dataset & MSPM Dataset~\cite{niu2023full,speth2024mspm} \\ 
\noalign{\smallskip} \hline \noalign{\smallskip}
\#Subjects & 1094 & 103\\
\#Data sets & 2812 & 205 \\
Age range & 18-91 & 18-58 \\
SBP range & 88-234 & 96-156  \\
DBP range & 56-148 & 52-106  \\
Mean SBP & 127.79 & 117.00  \\
Mean DBP & 85.40 & 75.34  \\
STD SBP & 18.26  & 11.27  \\
STD DBP & 12.47 & 8.84 \\
\noalign{\smallskip} \noalign{\hrule height 1.3pt} \\
\end{tabular}
} \vspace{-10pt}
\end{center}
\end{table}
\setlength{\tabcolsep}{1.4pt}

\subsection{Dataset}
\label{sec:dataset}
\begin{figure}[t]
\begin{center}
\includegraphics[scale=0.28]{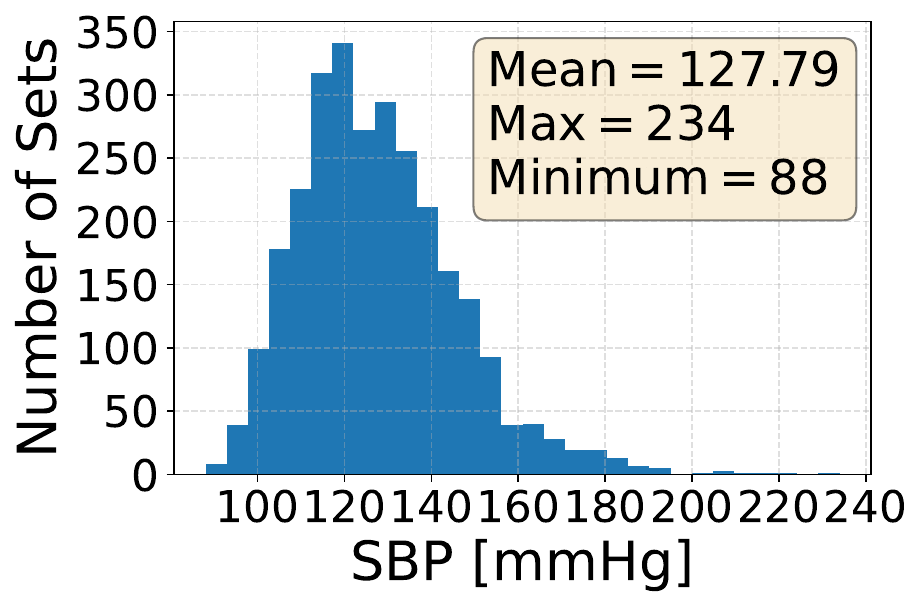}
%\hspace{-5pt}
\includegraphics[scale=0.28]{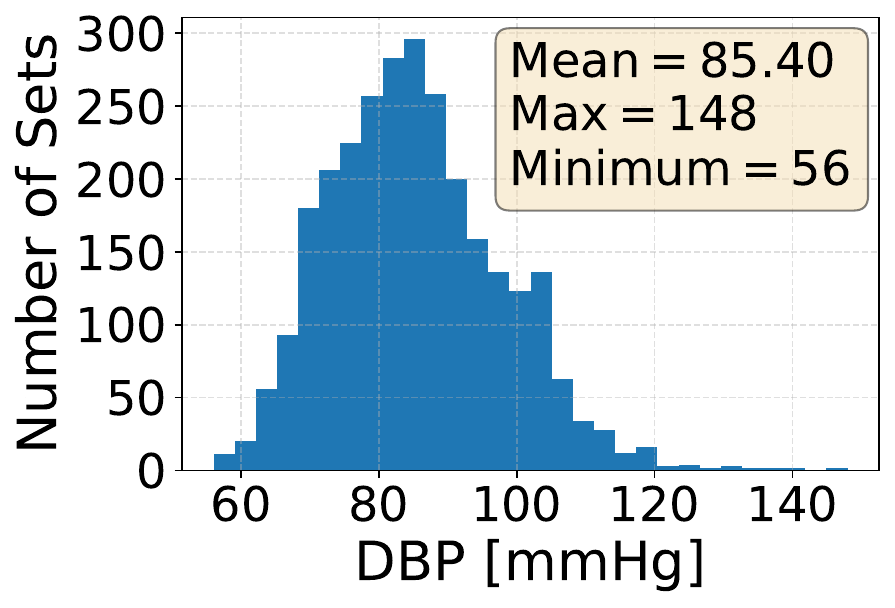}
\end{center}
\vspace{-12pt}
\caption{SBP and DBP distributions in FaceBP dataset.}
\vspace{-10pt}
\label{fig:dist}
\end{figure}

{\bf FaceBP Dataset:}
We collected a large dataset containing face videos, BP measurements, and PPG signals.  
All procedures in this study were approved by the NEC Ethical Review Committee for the Life Sciences (approval number: LS2024-006).
Informed consent was obtained from all subjects who participated in this study.
The dataset includes 1094 subjects (White, Asian, and Black, ages 18–91, 626 males, 468 females; see Appendix~B for details), comparable in scale to previous large datasets~\cite{luo2019smartphone,wu2022facial}.  
Each subject's face video was recorded using a Logitech C920n webcam at 30fps for one minute while seated.  
Simultaneously, BP was measured using an A\&D UA-1200BLE electronic sphygmomanometer, and PPG signals were acquired from a Plux blood volume pulse sensor attached to the fingertip.  
Each subject participated in two to five recording sessions, with some contributing data over multiple days.  
The dataset comprises 2812 recording sets and covers a wide BP range (see Fig.~\ref{fig:dist}).  
For some recordings (1699 out of 2812), face videos were captured not only using the Logitech C920n webcam but also using the Google Pixel~7 and Apple iPhone 14.
The main text reports the experimental results using the webcam, while Appendix~C reports the differences in estimation performance among the three cameras.
\\
\vspace{-10pt}
\\
{\bf MSPM Dataset~\cite{niu2023full,speth2024mspm}:}
This publicly available dataset includes face videos, BP measurements, and PPG signals collected from 103 subjects (White, Asian, and Black, ages 18–58, 39.8\% male, 58.3\% female). 
We used frontal videos taken with a DFK 33UX290 RGB camera for about 50 seconds during two resting BP sessions.
The total number of videos is 205 sets, with an SBP range of 96–156 mmHg and a DBP range of 52–106 mmHg. 

The statistics for the FaceBP and MSPM datasets are shown in TABLE~\ref{table:stats}. 
The ranges of SBP and DBP in FaceBP dataset are broad, indicating that these ranges are comparable to or broader than those in previous large-scale datasets~\cite{luo2019smartphone,wu2022facial}. 
Luo et al.~\cite{luo2019smartphone} included 1328 subjects, but only those with SBP between 100 and 129 mmHg. 
Wu et al.~\cite{wu2022facial} included 1143 subjects across three datasets, with SBP between 78 and 224 mmHg and DBP between 42 and 129 mmHg.

\subsection{Experimental Setup}
\label{sec:ex-setup}
We conduct 5-fold cross-validation in which the data of 80\% of the subjects are used as training data and 20\% as test data.
This setup ensures that the subjects do not overlap between the training and test data.
We use 20\% of the training data as validation data for monitoring loss values and tuning hyperparameters.
In each epoch, samples of 5 seconds (number of frames $F=150$, or a middle frame for face images) are randomly sampled and used for training the model.
We sample in the same manner during testing, and use the average of these estimated results.
The hyperparameters of the loss function $\mathcal{L}_{\rm pulse}$ are set to $\alpha=5$, $\beta=10$, and $\gamma=15$.
The sampling number for MC dropout is set to $T = 10$.
For BP estimation from rPPG and PPG signals, to prevent the estimated BPs from being concentrated around the mean BP, the training data are doubly oversampled when the SBP is below 110 or above 150 mmHg, or the DBP is below 70 or above 100 mmHg.
For BP estimation from face images, the images are resized to $H=224$ and $W=224$, and random crop, horizontal flip, color jitter, and grayscale are used as data augmentation.
The BP values are normalized to mean 0 and standard deviation 1.
For all models, we train for 30 epochs and use the model at the epoch with the lowest validation loss for testing.
The optimization algorithm is Adam~\cite{kingma2014adam}, batch size is 128, and learning rates are 1e-3, 1e-3, and 1e-4 for BP estimation from rPPG, PPG signals, and face images, respectively.
For the small-scale MSPM dataset, all models (including the proposed and comparative methods) are pre-trained on the large-scale FaceBP dataset and subsequently fine-tuned on MSPM using its face videos and corresponding BP labels.
Our models are implemented with PyTorch~\cite{paszke2019pytorch} and trained on eight A100 or L40S GPUs.
The training and inference times are short, and the inference can run in about two seconds even on a CPU environment (see Appendix~D).

\setlength{\tabcolsep}{4pt}
\begin{table*}[t]
\caption{Performance comparison using FaceBP and MSPM datasets. Results are based on 5-fold cross-validation, where training and test subjects are not overlapping. Estimation performance improves when evaluation metric given $\downarrow$ is lower and evaluation metric given $\uparrow$ is higher. Best results are marked in \textbf{bold} and second best in \underline{underline}.}
\vspace{-10pt}
\label{table:compare-bp}
\begin{center}
\scalebox{1}{
\begin{tabular}{lccccccccccccccc}  \noalign{\hrule height 1.3pt} \noalign{\smallskip}
        & \multicolumn{7}{c}{FaceBP Dataset} & &  \multicolumn{7}{c}{MSPM Dataset~\cite{niu2023full,speth2024mspm}}  
        \\ \noalign{\smallskip}
        \cline{2-8} \cline{10-16} \noalign{\smallskip}
        & \multicolumn{3}{c}{SBP} & &  \multicolumn{3}{c}{DBP} & & \multicolumn{3}{c}{SBP} & &  \multicolumn{3}{c}{DBP}
        \\ \noalign{\smallskip}
        \cline{2-4} \cline{6-8} \cline{10-12} \cline{14-16}
        \noalign{\smallskip}
        Method &  MAE$\downarrow$ & Corr.$\uparrow$ & Suc10$\uparrow$ & & MAE$\downarrow$ & Corr.$\uparrow$ & Suc10$\uparrow$ & & MAE$\downarrow$ & Corr.$\uparrow$ & Suc10$\uparrow$ & & MAE$\downarrow$ & Corr.$\uparrow$ & Suc10$\uparrow$   \\ \noalign{\smallskip} \hline
        \noalign{\smallskip}
        Mean Regressor & 14.29 & -0.035 & 42.0 & & 9.85 & -0.016 & 57.1 & & 9.10 & -0.214 & 60.5 & & 7.22 & -0.261 & 74.1   \\
        Zhou et al.~\cite{zhou2019noninvasive} & 13.37 & 0.320 & 47.0 & & 9.47 & 0.291 & 59.4 & & 9.71 & 0.097 & 60.5 & & 7.68 & 0.011 & 69.3    \\
        ResNet1D~\cite{schrumpf2021assessment} & 12.59 & 0.473 & 47.7 & & 8.86 & 0.451 & 63.8 & & 9.10 & 0.128 & 58.0 & & \underline{6.45} & 0.344 & 75.6   \\
        AlexNet1D~\cite{schrumpf2021assessment} & 12.39 & 0.464 & 49.3 & & 8.71 & 0.450 & 64.1 & & 9.19 & 0.103 & 61.5 & & 6.58 & 0.340 & \underline{77.6}    \\
        S-Net~\cite{wu2022facial} & 12.16 & 0.497 & 49.4 & & \underline{8.57} & 0.488 & \underline{66.0} & & \underline{8.56} & \underline{0.304} & \underline{65.4} & & 6.67 & 0.370 & 76.6   \\
        FS-Net~\cite{wu2022facial} & \underline{11.72} & \underline{0.545} & \underline{51.8} & & 8.62 & \underline{0.492} & 65.2 & & 8.76 & 0.274 & 58.0 & & 6.52 & \underline{0.438} & 76.6     \\
        CWT-CNN~\cite{trirongjitmoah2024assessing} & 13.66 & 0.272 & 44.6 & & 9.29 & 0.324 & 60.7 & & 9.48 & -0.042 & 59.0 & & 7.10 & 0.136 & 72.2    \\
        Liu et al.\cite{liu2024ensemble} & 12.13 & 0.470 & 51.7 & & 8.69 & 0.453 & 65.3 & & 8.60 & 0.274 & 64.4 & & 6.96 & 0.149 & 75.1  \\
        {\bf U-FaceBP (Ours)} & {\bf 11.05} & {\bf 0.620} & {\bf 54.8} & & {\bf 7.90} & {\bf 0.597} & {\bf 69.0} & & {\bf 7.87} & {\bf 0.547} & {\bf 65.9} & & {\bf 6.28} & {\bf 0.501} & {\bf 82.9}
        \\ \noalign{\smallskip} \noalign{\hrule height 1.3pt}
        \end{tabular}
}
\vspace{-10pt}
\end{center}
\end{table*}
\setlength{\tabcolsep}{1.4pt}

\subsection{Main Results}
{\bf Performance Comparison with SoTA:} We first compare our U-FaceBP with the eight benchmark and SoTA BP estimation methods listed in Table~\ref{table:compare-bp} using FaceBP and MSPM datasets.
The Mean Regressor baseline always predicts the average SBP and DBP values from the training set.
ResNet1D~\cite{schrumpf2021assessment}, AlexNet1D~\cite{schrumpf2021assessment}, S-Net~\cite{wu2022facial}, and CWT-CNN~\cite{trirongjitmoah2024assessing} estimate BP from rPPG signals.
Zhou et al.~\cite{zhou2019noninvasive} and FS-Net~\cite{wu2022facial} also utilize rPPG signals, along with additional features such as BMI estimated from face images.
Liu et al.~\cite{liu2024ensemble} is an end-to-end BP estimation method from RGB frames of face videos, which won the first place in the face video-based BP estimation task in the 3rd RePSS challenge~\cite{sun20243rd}.
Implementation details for all comparative and proposed methods are provided in Appendix~E.
We use mean absolute error (MAE), Pearson correlation coefficient (Corr.), and the percentage of predictions with absolute error below 10 mmHg (Suc10), which were commonly used in previous studies~\cite{wu2022facial,li2022hybrid}, as the evaluation metrics.

\begin{figure}[t]
\begin{center}
\includegraphics[scale=0.29]{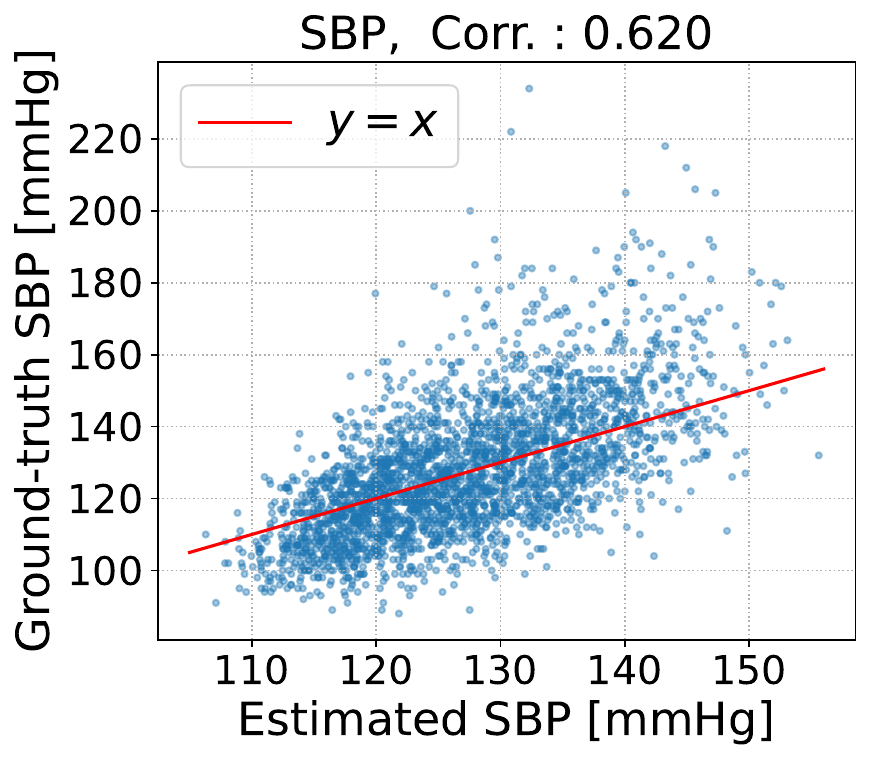}
%\hspace{-5pt}
\includegraphics[scale=0.29]{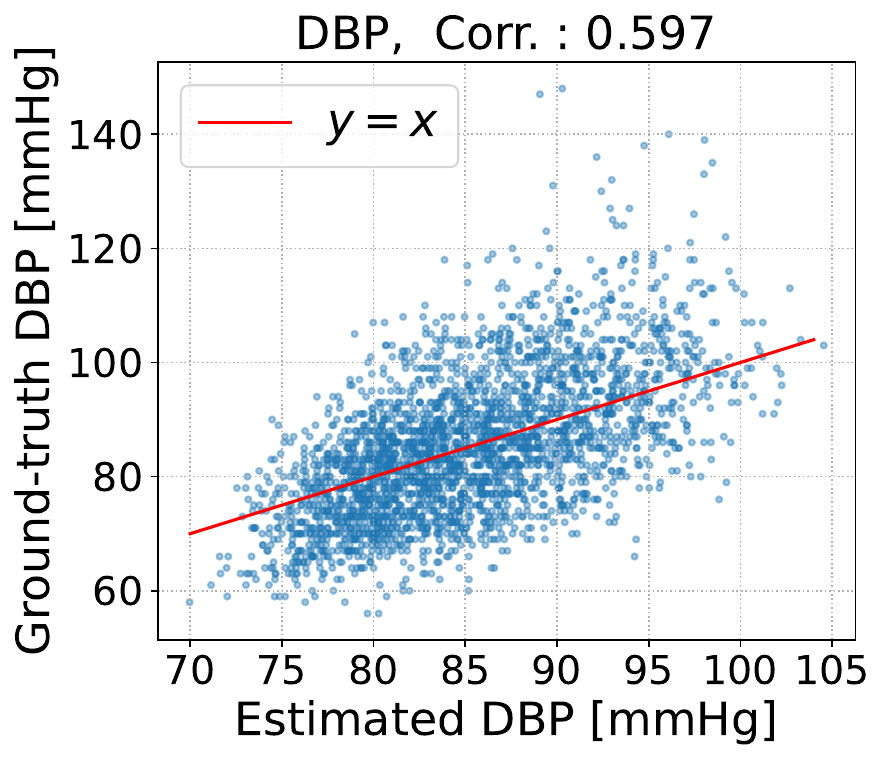}
\end{center}
\vspace{-12pt}
\caption{Correlation plots of SBP and DBP in U-FaceBP.}
\vspace{-10pt}
\label{fig:corr}
\end{figure}

Results for the FaceBP dataset are summarized in Table~\ref{table:compare-bp}.
The Mean Regressor baseline yields MAEs of 14.29 for SBP and 9.85 for DBP, with negligible correlation (Corr. close to zero).
Several methods perform only marginally better than this baseline, consistent with previous findings~\cite{schrumpf2021assessment,wu2022facial}.
In contrast, our U-FaceBP outperforms all comparative methods across all metrics, achieving MAEs of 11.05 and 7.90 for SBP and DBP, respectively.
Wilcoxon signed-rank tests conducted on the absolute errors between U-FaceBP and each comparative method confirm that the differences are statistically significant ($p < 0.01$), with Bonferroni correction applied for multiple comparisons.
Figure~\ref{fig:corr} shows the correlation plots for SBP and DBP predictions by U-FaceBP.
The resulting correlation coefficients are 0.620 for SBP and 0.597 for DBP, indicating a moderately strong positive relationship between the estimated and ground-truth values.

The MSPM dataset exhibits a narrower BP distribution, leading to generally smaller prediction errors across all methods.
Even under this condition, U-FaceBP consistently outperforms other approaches.
In the British Hypertension Society (BHS) standard~\cite{o1993british} using Suc10~\footnote[5]{To simplify evaluation, we use only Suc10 rather than including all of Suc5, Suc10, and Suc15.}, which is the medical criteria for BP measurement devices, U-FaceBP achieves Grade C (Suc10 $\geq$ 65\%) for DBP on FaceBP and SBP on MSPM, and Grade B (Suc10 $\geq$ 75\%) for DBP on MSPM.
Validated commercial cuff-based BP monitors typically achieve Grade A (Suc10 $\geq$ 85\%) or Grade B under the BHS standard~\cite{o2001blood}.
Although our method does not yet reach the validation grades achieved by cuff-based BP monitors, it demonstrates competitive performance in the context of emerging cuffless and contactless BP estimation approaches.
\\
\vspace{-5pt}
\\
{\bf Inter-dataset Performance Comparison:} We compare our BP estimation performance with that reported in four prior studies listed in Table~\ref{table:inter-bp}.
The first three studies estimate BP from face videos, while the last one estimates from contact-based PPG signals.
Since BP ranges differ across datasets, we adopt the mean absolute scaled error (MASE), proposed by Gonz{\'a}lez et al.~\cite{gonzalez2023benchmark}, to enable fair performance comparisons.
MASE is defined as the ratio of a model’s MAE to that of a mean regressor baseline, and we report the best MASE from each study.
As shown in Table~\ref{table:inter-bp}, our method outperforms previous face video-based BP estimation methods.
Notably, our U-FaceBP achieves a lower MASE than Gonz{\'a}lez et al.~\cite{gonzalez2023benchmark}, who conducted a comprehensive evaluation of BP estimation from PPG signals.
This result highlights that our BP estimation from face videos can achieve lower error than not only prior face video-based approaches, but also contact-based PPG estimation methods.
\\
\setlength{\tabcolsep}{4pt}
\begin{table}[t]
\caption{Inter-dataset performance comparison. Performance improves when MASE is lower.}
\vspace{-12pt}
\label{table:inter-bp}
\begin{center}
\scalebox{1}{
\begin{tabular}{cccccc}  \noalign{\hrule height 1.3pt} \noalign{\smallskip}
& & \multicolumn{1}{c}{SBP} & & \multicolumn{1}{c}{DBP}
\\ \noalign{\smallskip}
\cline{3-3} \cline{5-5}
\noalign{\smallskip}
Study & Input & MASE$\downarrow$ & & MASE$\downarrow$  \\ \noalign{\smallskip} \hline
\noalign{\smallskip}
Schrumpf et al.~\cite{schrumpf2021assessment} & Face video & 97.2 & & 93.0 \\
Wu et al.~\cite{wu2022facial} & Face video & 86.8 & & 93.4 \\
Wu et al.~\cite{wu2022camera} & Face video & 81.2 & & 89.2\\
Gonz{\'a}lez et al.~\cite{gonzalez2023benchmark}& PPG signals* & 79.8 & & 90.2 \\
{\bf Ours} & Face video & {\bf 77.3} & & {\bf 80.2}
\\ \noalign{\smallskip} \noalign{\hrule height 1.3pt}
*acquired from PPG sensor
\end{tabular}
}
\vspace{-10pt}
\end{center}
\end{table}
\setlength{\tabcolsep}{1.4pt}
\vspace{-5pt}
\\
{\bf Cross-dataset Estimation Performance:}
We evaluate the cross-dataset generalization performance by training U-FaceBP on the large-scale FaceBP dataset and testing it on the small-scale MSPM dataset.
We consider four variations.
\begin{itemize}
\item Zero-shot (FaceBP $\to$ MSPM): The model trained on FaceBP is directly tested on MSPM.
\item Zero-shot + Norm (FaceBP $\to$ MSPM): The model trained on FaceBP is tested on MSPM, but the mean and standard deviation from MSPM (excluding test data) are used to scale back the normalized BP predictions to the original scale.
\item MSPM Only: The model is trained and tested on MSPM using subject-independent 5-fold cross-validation.
\item FaceBP Pretrain + MSPM Fine-tune (final setup): The model is pre-trained on FaceBP and fine-tuned on MSPM, and tested on MSPM using the same cross-validation.
\end{itemize}
TABLE~\ref{table:cross} shows cross-dataset estimation performance in four variations.
Zero-shot exhibits poor MAE, which may be attributed to the statistical differences between the FaceBP and MSPM datasets (see the mean and standard deviation (STD) in TABLE~\ref{table:stats}).
Interestingly, Zero-shot achieves the highest correlation performance for DBP, suggesting that relative ranking information is partially preserved across datasets.
Zero-shot + Norm, which aligns the prediction scale with MSPM statistics, significantly reduces MAE and achieves performance comparable to MSPM Only.
Finally, FaceBP Pretrain + MSPM Fine-tune achieves the best overall performance, demonstrating the effectiveness of transfer learning from the large-scale FaceBP dataset to the smaller MSPM dataset.

\setlength{\tabcolsep}{4pt}
\begin{table}[t]
\caption{Cross-dataset estimation performance on MSPM dataset.}
\vspace{-15pt}
\label{table:cross}
\begin{center}
\scalebox{0.95}{
        \begin{tabular}{lcccccc}  \noalign{\hrule height 1.3pt} \noalign{\smallskip}
        &  \multicolumn{5}{c}{MSPM Dataset}  
        \\ \noalign{\smallskip}
        \cline{2-6} \noalign{\smallskip}
        & \multicolumn{2}{c}{SBP} & &  \multicolumn{2}{c}{DBP}
        \\ \noalign{\smallskip}
        \cline{2-3} \cline{5-6}
        \noalign{\smallskip}
        U-FaceBP &  MAE$\downarrow$ & Corr.$\uparrow$ & & MAE$\downarrow$ & Corr.$\uparrow$    \\ \noalign{\smallskip} \hline
        \noalign{\smallskip}
        Zero-shot (FaceBP $\to$ MSPM) & 11.49 & 0.261 & & 9.74 & {\bf 0.529}\\
        Zero-shot + Norm (FaceBP $\to$ MSPM) & 9.01 & 0.215 & & 6.58 & 0.459 \\
        MSPM Only & 8.37 & 0.389 & & 6.91 & 0.203  \\
        {\bf FaceBP Pretrain + MSPM Fine-tune} & {\bf 7.87} & {\bf 0.547} & & {\bf 6.28} & 0.501
        \\ \noalign{\smallskip} \noalign{\hrule height 1.3pt}
        \end{tabular}
        }
\end{center}
\end{table}
\setlength{\tabcolsep}{1.4pt}

\subsection{Ablation Study}
\begin{figure}[t]
\begin{center}
\includegraphics[scale=0.29]{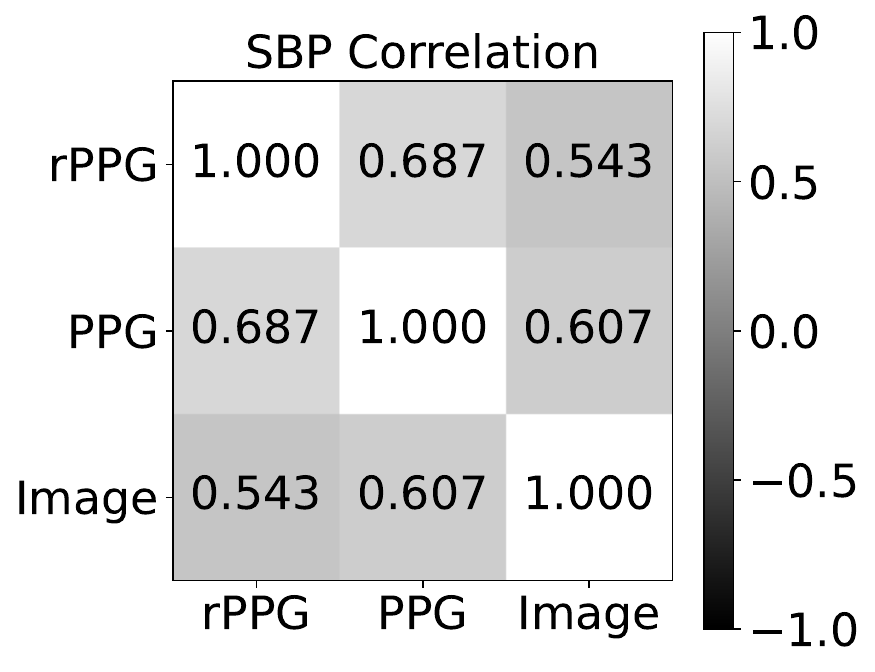}
%\hspace{-5pt}
\includegraphics[scale=0.29]{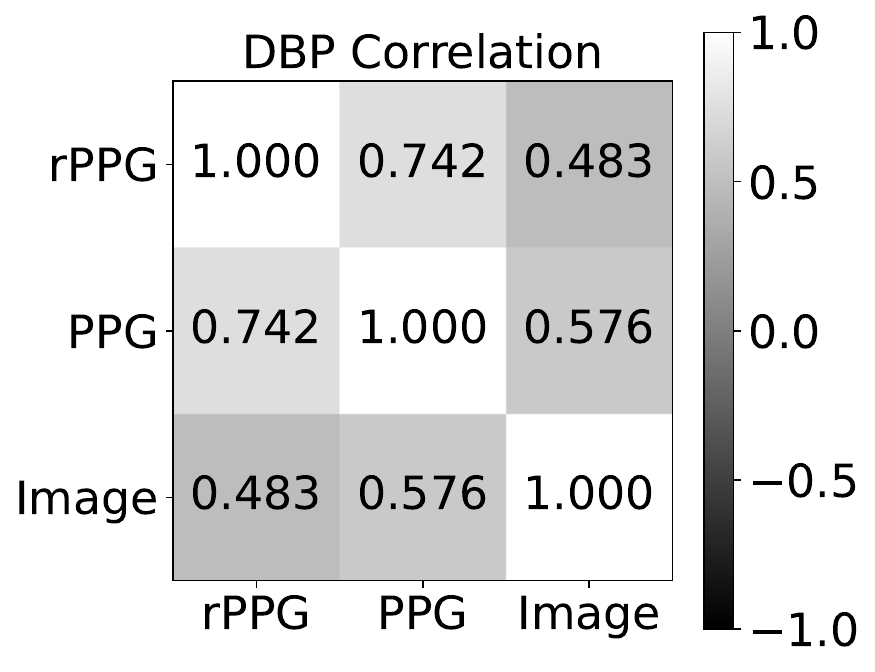}
\end{center}
\vspace{-12pt}
\caption{Correlation coefficients of predicted SBP and DBP among ensemble components on FaceBP dataset.}
\vspace{-10pt}
\label{fig:corr_heat}
\end{figure}

We ablate the components of U-FaceBP using the FaceBP dataset to verify the effectiveness of the following factors.
\\
{\bf Ensemble Components of U-FaceBP:}
Table~\ref{table:ablation} (a-g) shows the performance comparison among the ensemble components of U-FaceBP.
The BP estimation models using only rPPG, PPG signals, or face images (Image) perform well individually, with PPG achieving overall better results.
This superior performance of PPG demonstrates the advantage of extracting accurate pulse waveforms from face videos.
The performance of ``rPPG+PPG" is better than that of each individual component, and the incorporation of face images further improves the performance of rPPG or PPG signals.
These results suggest that each of the three components provides complementary information related to BP.
Notably, BP estimation based on pulse waves outperforms that based on face images, as shown by ``rPPG+PPG" vs. ``Image".
This is because facial appearance (Image) captures overall BP tendencies, whereas the pulse wave (rPPG+PPG) directly reflects hemodynamic differences.
Ultimately, ``rPPG+PPG+Image" exhibits the best performance.
\\
Figure~\ref{fig:corr_heat} shows the correlation coefficients of the predicted BP among ensemble components.
The rPPG and PPG components share the same BP estimation mechanism—pulse waveform analysis—resulting in high prediction correlations (0.687 for SBP and 0.742 for DBP).
On the other hand, rPPG/PPG and images have different BP prediction processes—physiological and visual features, respectively—resulting in moderate predictive correlations.
Consequently, since each component provides complementary predictions, their ensemble enhances estimation performance.
\setlength{\tabcolsep}{4pt}
\begin{table}[t]
\caption{Ablation study for U-FaceBP using FaceBP dataset. Shading indicates our final setting.}
\vspace{-15pt}
\label{table:ablation}
\begin{center}
\scalebox{1}{
\begin{tabular}{lccccc}  \noalign{\hrule height 1.3pt} \noalign{\smallskip}
& \multicolumn{2}{c}{SBP} & & \multicolumn{2}{c}{DBP}
\\ \noalign{\smallskip}
\cline{2-3} \cline{5-6}
\noalign{\smallskip}
Ablation &  MAE$\downarrow$ & Corr.$\uparrow$ & & MAE$\downarrow$  & Corr.$\uparrow$   \\ \noalign{\smallskip} \hline
\noalign{\smallskip}
(a) rPPG & 11.95 & 0.528 & & 8.46 & 0.505    \\
(b) PPG** & 11.77 & 0.553 & & 8.33 & 0.540    \\
(c) Image & 11.78 & 0.518 & & 8.55 & 0.486   \\
(d) rPPG+PPG** & 11.42 & 0.588 & & 8.14 & 0.560    \\
(e) rPPG+Image & 11.19 & 0.596 & & 7.99 & 0.575  \\
(f) PPG**+Image & 11.11 & 0.601 & & 8.00 & 0.578    \\
\rowcolor[rgb]{0.9, 0.9, 0.9} (g)
rPPG+PPG**+Image & 11.05 & 0.620 & & 7.90 & 0.597 \\
(h) Late Fusion &12.37 & 0.464 & & 9.07 & 0.434 \\
(i) rPPG+PPG**+Image w/o UAN & 11.75 & 0.580 & & 8.16 & 0.571  \\
(j) PPG** w/o VPG**+APG** & 12.07 & 0.510 & & 8.39 & 0.521   \\
(k) Image w/o FRL  & 12.18 & 0.481 & & 8.87 & 0.437
\\ \noalign{\smallskip} \noalign{\hrule height 1.3pt}
**estimated from face videos
\end{tabular}
}
\vspace{-5pt}
\end{center}
\end{table}
\setlength{\tabcolsep}{1.4pt}
\\
{\bf Ensemble vs. Late Fusion:}
Although we use an ensemble approach using rPPG, PPG signals, and face images, the feature fusion of these components is a straightforward approach.
Table~\ref{table:ablation} (h) indicates the BP estimation performance of the single BNN model that concatenates the feature representations of the three components using late fusion (the number of trainable parameters for ensemble and late fusion are comparable).
Interestingly, the late fusion model significantly degrades performance compared with the ensemble approach in Table~\ref{table:ablation} (g).
This may be because if one modality makes an inaccurate prediction, it negatively affects the late fusion model~\cite{swati2024exploring}.
On the other hand, ensemble methods are not easily affected by a single modality, resulting in a superior performance to late fusion.
\\
{\bf Uncertainty-aware Network:}
Table~\ref{table:ablation} (g,i) compares the performance of an uncertainty-aware network (UAN) with a standard NN using mean aggregation, where both models have a comparable number of trainable parameters.  
The results show that incorporating BNNs into the three ensemble components and aggregating them using UDA enhances performance.  
This suggests that BNNs and UDA effectively mitigate overfitting when learning from small and noisy datasets and enable more reliable aggregation by considering uncertainty (see Appendix~F for details on UDA weights).  
\\
{\bf Derivatives of PPG Signal:}
Table~\ref{table:ablation} (b,j) shows the effectiveness of introducing derivatives of PPG signals ($i.e.$, VPG and APG signals in Fig.~\ref{fig:method} (b)).
The derivatives provide valuable information related to BP (aortic compliance and stiffness~\cite{liu2017cuffless}), leading to better performance.
\\
{\bf Facial Representation Learning (FRL):}
Table~\ref{table:ablation} (c,k) compares BP estimation performance from face images with and without FRL, where the latter uses ImageNet pre-training~\cite{deng2009imagenet}.  
The results show that leveraging FRL improves BP estimation performance compared to the commonly used ImageNet pre-training.

\begin{figure*}[t]
\begin{center}
\includegraphics[scale=0.36]{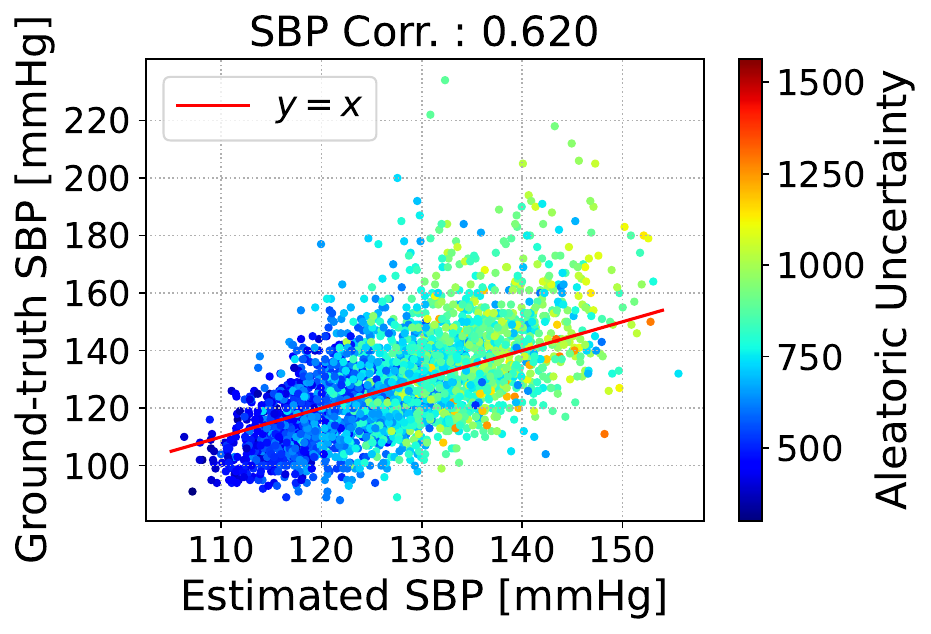}
\includegraphics[scale=0.36]{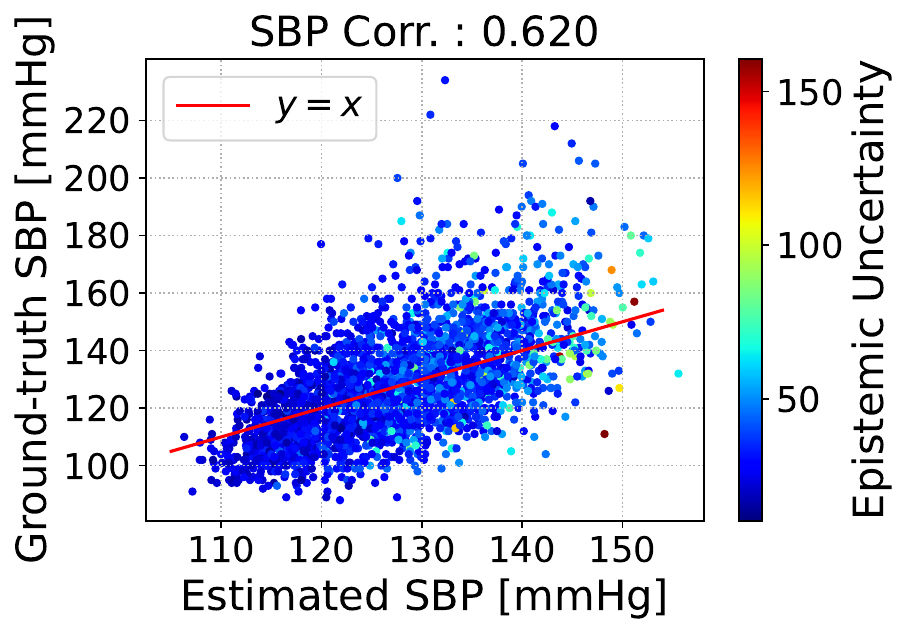}
\includegraphics[scale=0.36]{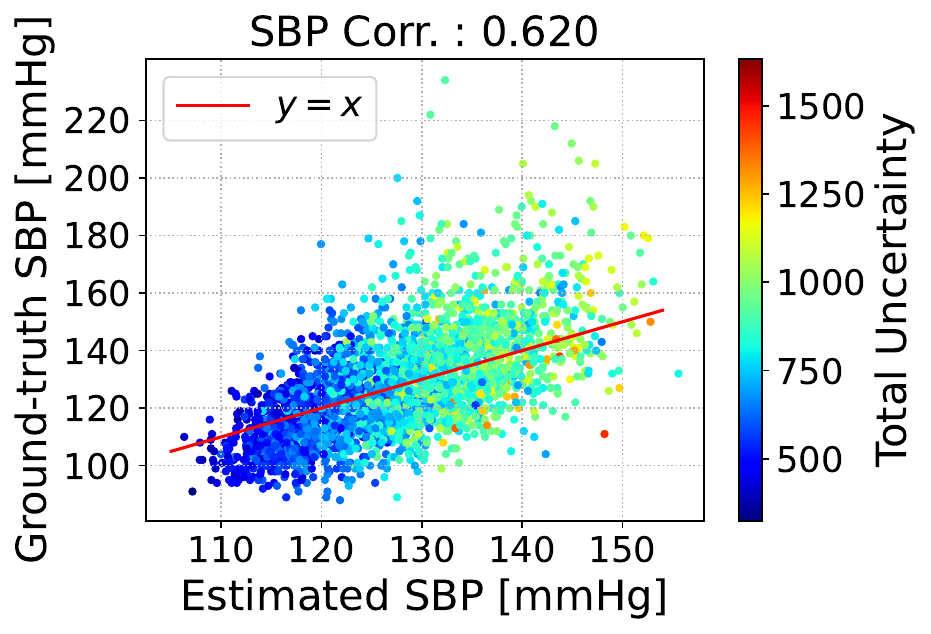}
\\
\hspace{-8pt}
\includegraphics[scale=0.36]{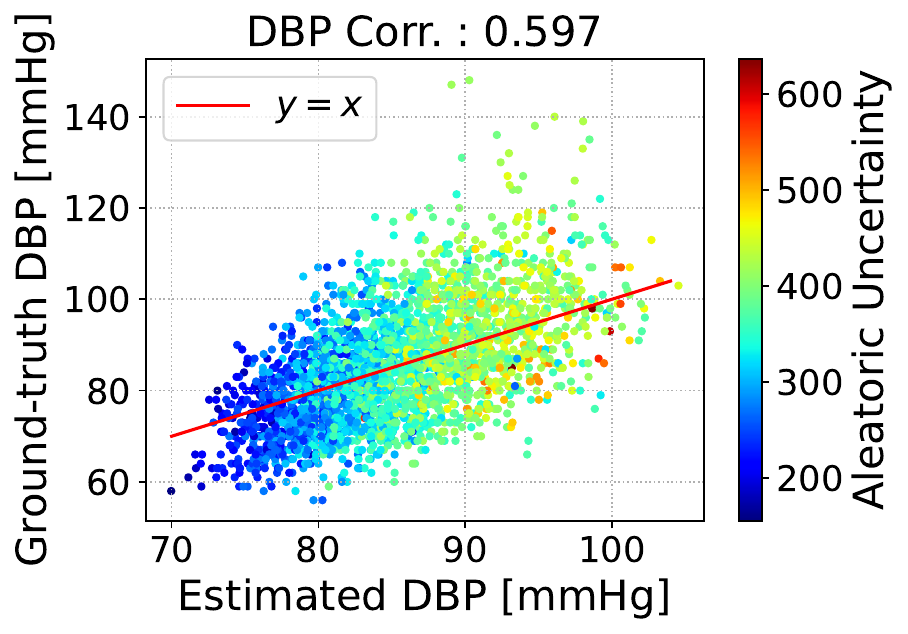}
\hspace{1pt}
\includegraphics[scale=0.36]{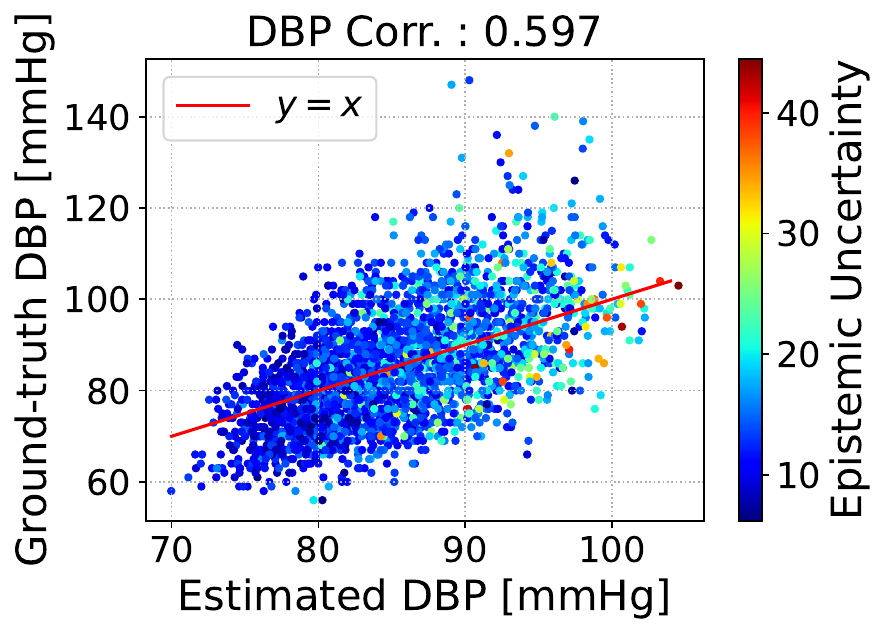}
\hspace{1pt}
\includegraphics[scale=0.36]{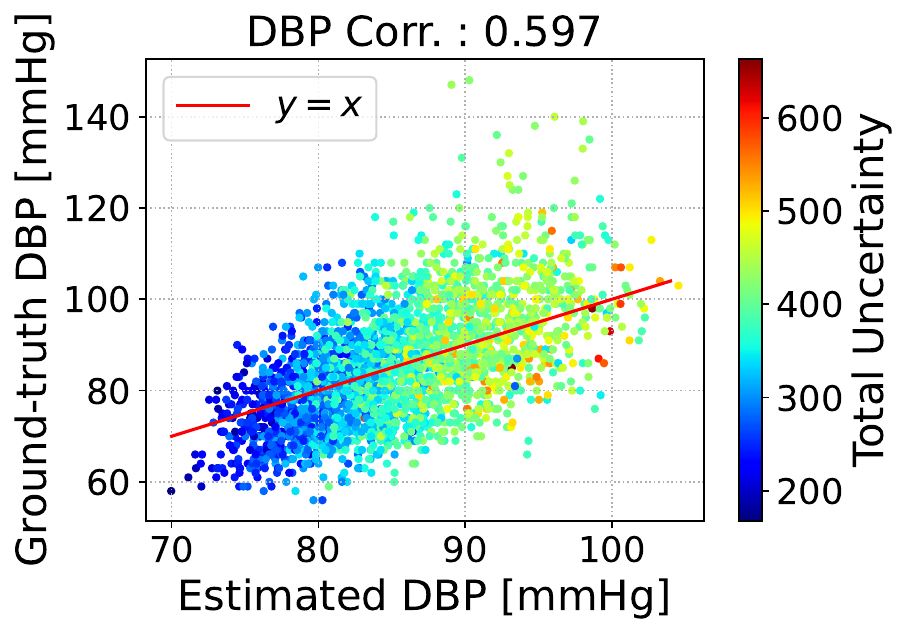}
\end{center}
\vspace{-8pt}
\caption{Correlation plots for SBP and DBP predictions by U-FaceBP with aleatoric, epistemic, and total uncertainties on FaceBP dataset.
First and second rows correspond to SBP and DBP predictions, respectively. First, second, and third columns represent aleatoric, epistemic, and total uncertainties, respectively.}
%\vspace{-5pt}
\label{fig:corr_total}
\end{figure*}

\begin{figure}[t]
\begin{center}
\includegraphics[scale=0.29]{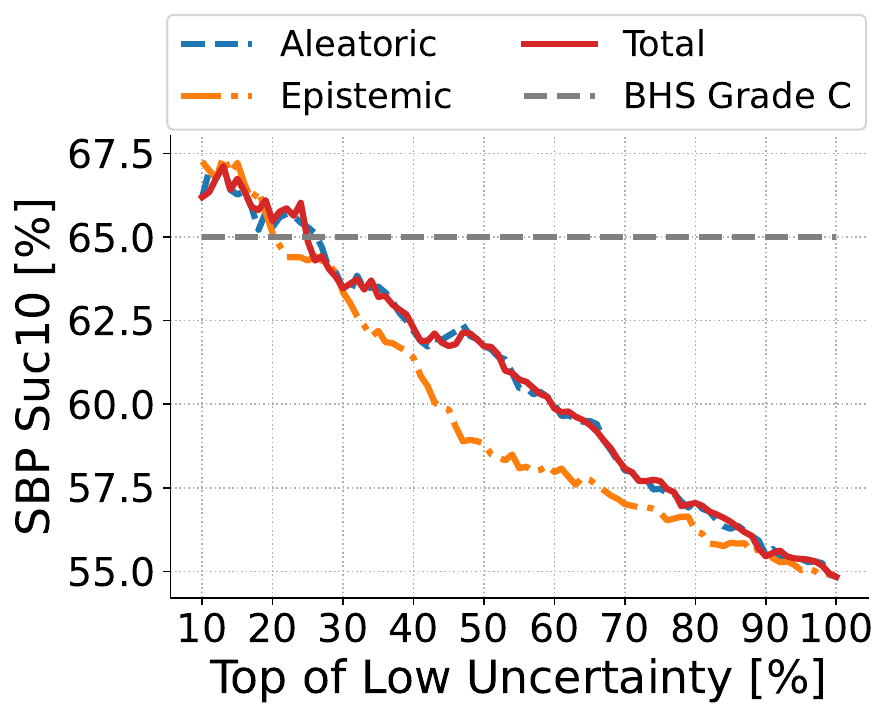}
%\hspace{1pt}
\includegraphics[scale=0.29]{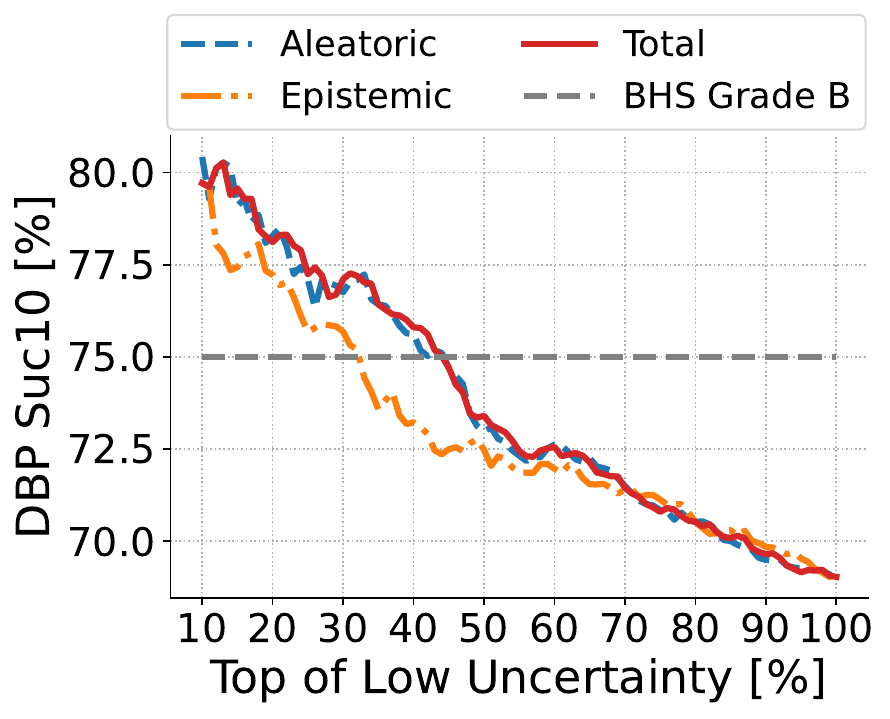}
\end{center}
\vspace{-10pt}
\caption{Relationship between top of low uncertainty (top $x$\% of predictions ranked in ascending order of uncertainty) and their Suc10.}
%\vspace{-5pt}
\label{fig:confidence}
\end{figure}

\subsection{Uncertainties}
\label{sec:unc_result}
We investigate uncertainties using the FaceBP dataset.
\\
{\bf Uncertainty Trends:} Figure~\ref{fig:corr_total} shows the correlation plots for SBP and DBP predictions by U-FaceBP with aleatoric, epistemic, and total uncertainties on the FaceBP dataset.
The aleatoric uncertainty tends to be larger when the estimated BP is higher, reflecting the fact that the error can be large for SBP above 130 mmHg and DBP above 80 mmHg in some cases.
This suggests that BP estimation is difficult when BP is high, $i.e.$, the predictive variance in NLL loss (see Eq.~(1)) is difficult to reduce.
This difficulty is associated with the fact that hypertension has a wide BP range ($e.g.$, from 140 to 234 mmHg for SBP in the FaceBP dataset).
The epistemic uncertainty remains relatively uniform across a broad range of estimated BP values ($e.g.$, from 110 to 140 mmHg for SBP), indicating that the training data is sufficient within this range (see the histogram in Fig.~\ref{fig:dist}).
In contrast, higher uncertainty is observed in the hypertensive range ($e.g.$, SBP $\geq$ 140 mmHg), suggesting a lack of training samples in this region. This elevated uncertainty accurately reflects the model’s lack of knowledge due to data sparsity.
The total uncertainty shows a similar trend, particularly influenced by aleatoric uncertainty, with higher estimated BP tending to result in higher uncertainty.
\begin{figure}[t]
\begin{center}
\includegraphics[scale=0.29]{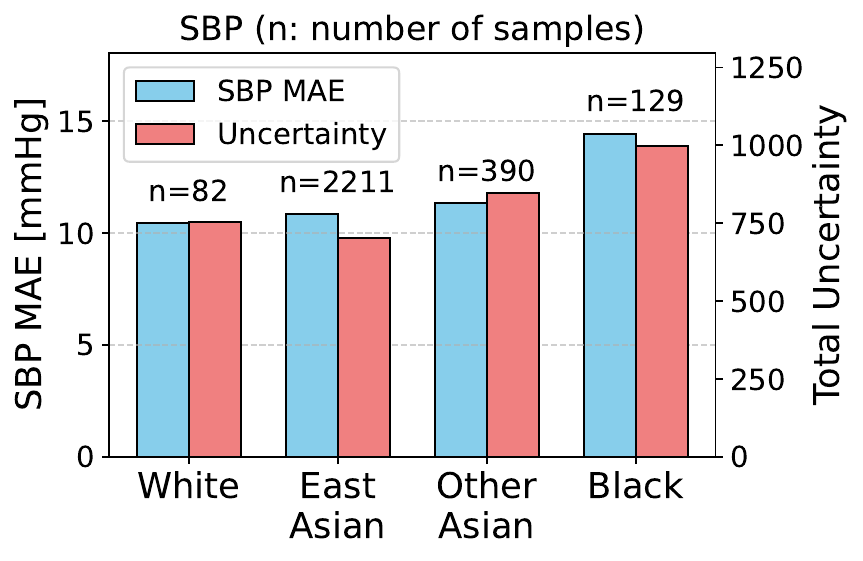}
%\hspace{1pt}
\includegraphics[scale=0.29]{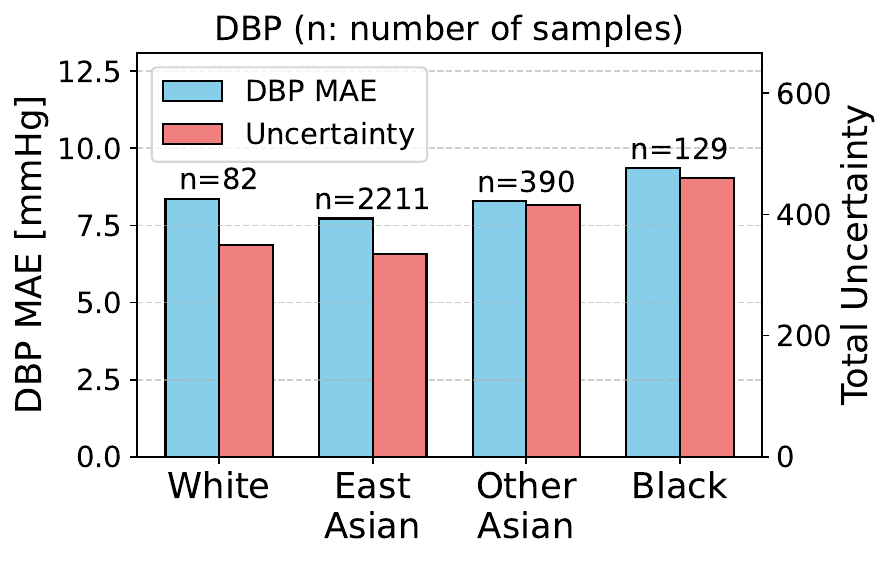}
\end{center}
\vspace{-10pt}
\caption{MAE and total uncertainty for each racial group. “Other Asian” includes Central, South, and Southeast Asian.}
%\vspace{-5pt}
\label{fig:unc_race}
\end{figure}
\\
{\bf Confidence Assessment:} Figure~\ref{fig:confidence} shows the relationship between uncertainties and estimation performance.
The horizontal axis represents the top $x$\% of predictions ranked in ascending order of uncertainty, referred to as the ``top of low uncertainty".
Suc10 increases as aleatoric and epistemic uncertainties decrease, with total uncertainty showing a similar trend.
The top 30–40\% of predictions with low uncertainty meet the BHS Grade B or C and mostly correspond to normotensive cases (see Fig.~\ref{fig:corr_total}), supporting a triage-like scenario, where clearly healthy cases can be handled by AI, while uncertain ones are directed to experts or BP devices.
\\
{\bf Racial Subgroup Analysis:} Figure~\ref{fig:unc_race} presents the MAE and total uncertainty for each racial group.
Compared to individuals with lighter skin tones such as White and East Asian, groups with darker skin tones such as Other Asian ($e.g.,$ South and Southeast Asian) and Black tend to exhibit slightly higher MAEs.
This is likely because darker skin, with higher melanin concentrations, absorbs more light and reduces the signal-to-noise ratio—a well-known challenge in prior rPPG studies~\cite{nowara2020meta}.
Notably, groups with darker skin tones show higher total uncertainty, suggesting that our uncertainty estimation framework implicitly captures BP prediction difficulty.

\begin{figure}[t]
\begin{center}
\includegraphics[scale=0.65]{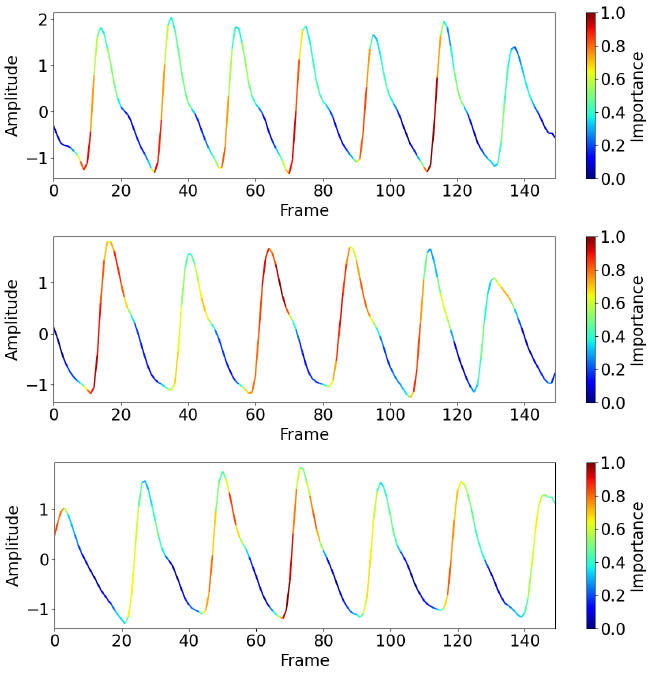}
\end{center}
\vspace{-10pt}
\caption{Reconstructed PPG signals and corresponding temporal importance.}
%\vspace{-5pt}
\label{fig:atten}
\end{figure}

\subsection{Visualization}
To improve the interpretability of U-FaceBP, we conduct visualization analyses to identify the temporal segments of pulse waveforms and the spatial regions of face images that most strongly influence BP prediction.
These analyses allow us to assess whether the model focuses on physiologically plausible features.

We first analyze the temporal segments emphasized by ResNet1D, which processes PPG signals reconstructed from facial videos.
Following Li et al.~\cite{li2022hybrid}, we extract feature maps from a convolutional layer after model training.
Specifically, considering the balance between higher-level feature representation and temporal resolution, we compute the absolute values of the feature maps from the third convolutional layer of ResNet1D and average them across all channels.
Figure~\ref{fig:atten} shows the reconstructed PPG signals from face videos and the corresponding temporal importance, which are calculated from the feature maps.
The results indicate that the rising portion of the pulse wave is most consistently important.
This pattern is also observed in subjects other than the three shown in Fig.~\ref{fig:atten}, and this finding is consistent with the results of Li et al.~\cite{li2022hybrid}.
The rising portion corresponds to slope transit time (STT), which has been shown to be associated with BP~\cite{addison2016slope}.
Therefore, these results suggest that the pulse wave segments emphasized by the model align with physiologically meaningful regions relevant to BP prediction.

\begin{figure}[t]
\begin{center}
\includegraphics[scale=0.4]{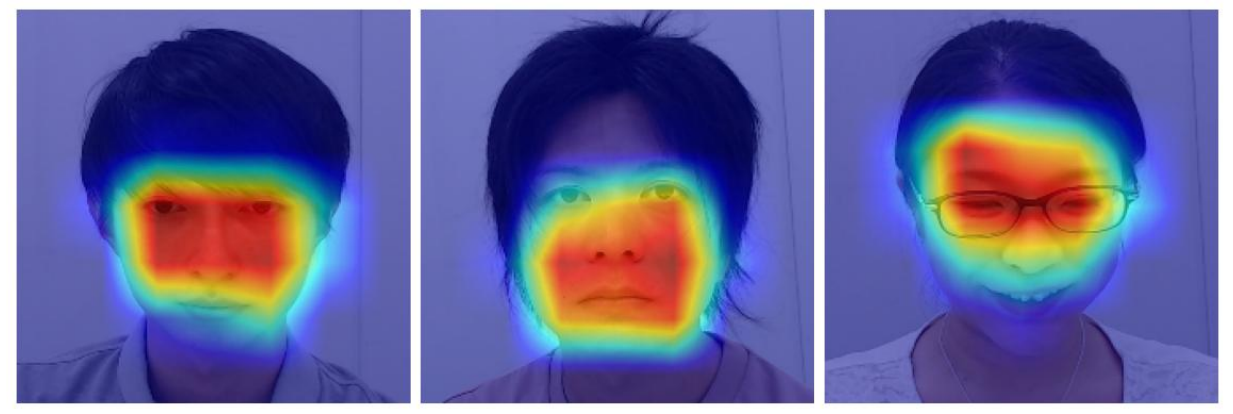}
\end{center}
\vspace{-10pt}
\caption{Saliency maps for predicting BP from face images.}
%\vspace{-5pt}
\label{fig:cam}
\end{figure}

We then analyze the spatial saliency maps of face images.
The saliency maps are obtained using Eigen-CAM~\cite{muhammad2020eigen} from the final block of ResNet50, which predicts BP from face images.
Figure~\ref{fig:cam} illustrates saliency maps for face images of three representative subjects.
The results show that the facial region—excluding hair, clothing, and background—is consistently emphasized across subjects. 
This region primarily includes facial skin, the hairline, and facial contours, which may reflect features such as wrinkles, baldness, and BMI associated with BP.
These results suggest that relevant facial characteristics associated with BP are emphasized in the prediction process.

\section{Conclusions and Limitations}
We propose U-FaceBP, an uncertainty-aware Bayesian ensemble deep learning method for face video-based BP estimation.
By incorporating uncertainty and a powerful ensemble approach, our U-FaceBP outperforms the SoTA methods and meets some of the medical criteria.
We also demonstrate that uncertainty is useful for modality fusion, confidence assessment, and racial subgroup analysis.
Face video-based BP estimation remains an emerging research area, and limitations persist, particularly in accurately estimating BP for severe hypertension cases  
(SBP$\ge$160 mmHg, DBP$\ge$105 mmHg).  
However, U-FaceBP successfully identifies 97.6\% (calculated from Fig.~\ref{fig:corr}) of hypertensive patients (SBP$\ge$140 mmHg) as having BP above the normal range (SBP$<$120 mmHg).
This suggests that U-FaceBP holds promise for BP monitoring, particularly for individuals who may not regularly track their health.

Regarding the limitations in estimation performance for severe hypertension, we acknowledge that the current dataset may lack sufficient cases of extreme hypertension.
Future work will include collecting additional cases of severe hypertension and hypotension in clinical settings to improve estimation performance in these ranges.
In addition, for real-time inference in mobile environments ($e.g.,$ smartphones and tablets), future work will explore lightweight architectures or knowledge distillation techniques.

\section*{Appendix A}
\label{app:ptt}
\begin{figure}[h]
\begin{center}
\includegraphics[scale=0.295]{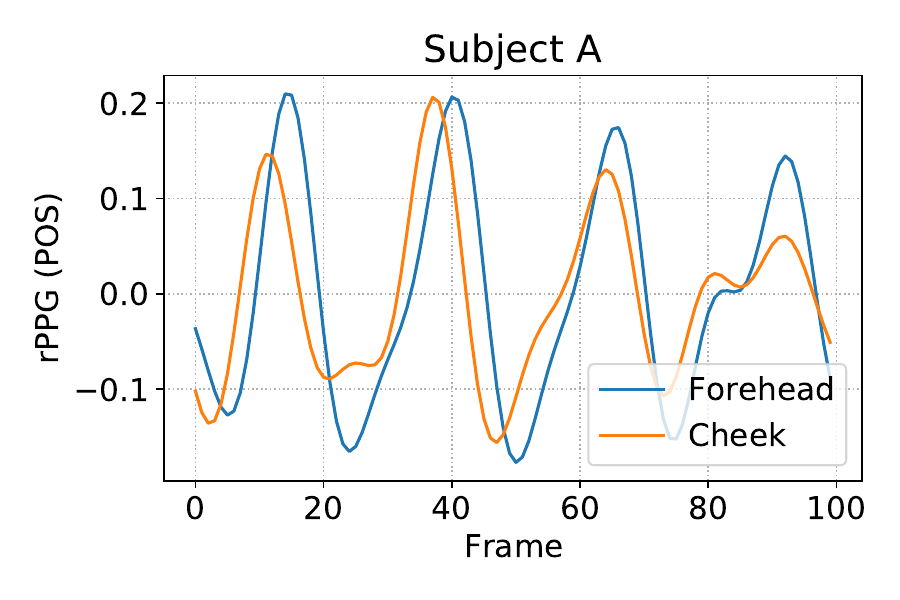}
\hspace{-10pt}
\includegraphics[scale=0.295]{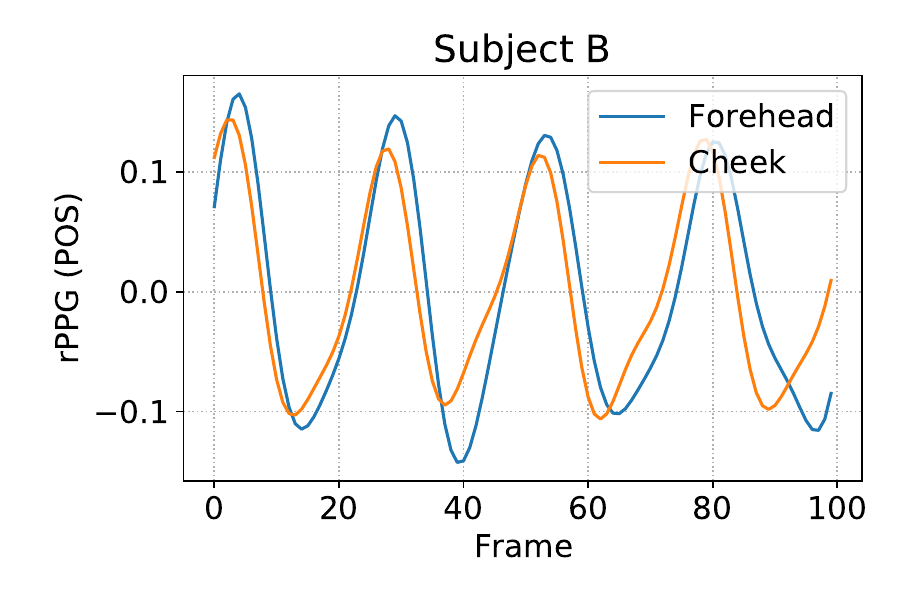}
\end{center}
\vspace{-10pt}
\caption{rPPG signals from cheek and forehead in FaceBP dataset.}
\vspace{-5pt}
\label{fig:ptt}
\end{figure}

It has been demonstrated in prior studies~\cite{wu2022facial,park2024robust} that there is a sufficient shift in the peak of rPPG signals at 30fps from different parts of face.
Figure~\ref{fig:ptt} shows that there are clear shifts of rPPG signals in our FaceBP dataset as well.
Note that the rPPG signals represent POS~\cite{wang2016algorithmic} after band-pass filtering.
Figure~\ref{fig:ptt} represents that pulse waves on the cheek are ahead of those on the forehead.
This result makes sense, since the heart pulse arrives later at locations further away from the heart.
However, we note that such clear phase differences are not consistently 
observed across all subjects. 
In some cases, the shifts are subtle due to inter-subject variability, signal-to-noise ratio differences, and physiological factors such as vascular elasticity.
In the same manner as Wu et al.~\cite{wu2022facial}, we input rPPG signals from three different parts into S-Net which extracts PTT from multiple rPPG signals.

\section*{Appendix B}
\label{app:data}
\begin{figure}[h]
\begin{center}
\includegraphics[scale=0.33]{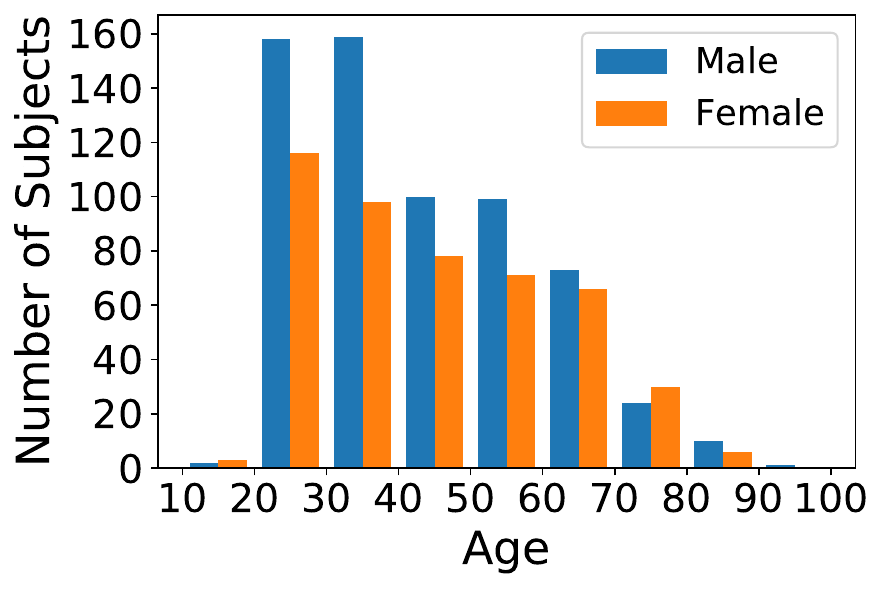}
\end{center}
\vspace{-10pt}
\caption{Gender and age distribution of subjects in FaceBP dataset.}
\vspace{-5pt}
\label{fig:age}
\end{figure}

\begin{figure}[h]
\begin{center}
\includegraphics[scale=0.35]{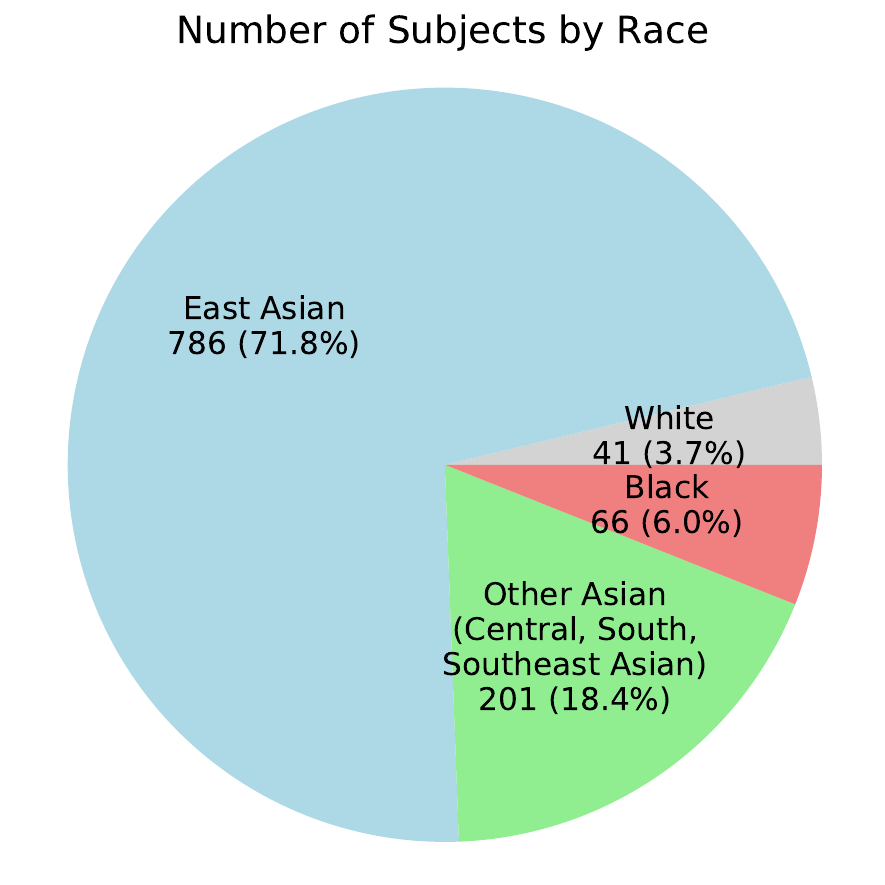}
\end{center}
\vspace{-13pt}
\caption{Number of subjects for each racial group in FaceBP dataset.}
\vspace{-5pt}
\label{fig:race}
\end{figure}

Figure~\ref{fig:age} shows the gender and age distribution of subjects in the FaceBP dataset.
The FaceBP dataset has a well-balanced gender distribution overall and includes a broad age range, spanning from individuals in their 10s to 90s.
Figure~\ref{fig:race} illustrates the number of subjects for each racial group in the FaceBP dataset. While the majority of subjects are East Asian, the dataset also includes a diverse set of racial groups, such as Other Asian (Central, South, and Southeast Asian), White, and Black.
Our racial grouping is a simplified version of the race categories defined in the FairFace dataset~\cite{karkkainen2021fairface}.

As mentioned in the dataset subsection, each subject participated in two to five recording sessions.
One of these sessions involved a BP-raising experiment, while the remaining sessions were conducted in a resting state.
In the BP-raising experiment, subjects placed their feet on a stool and applied pressure with the instep, following the protocol used in previous studies~\cite{wu2022facial,wu2022camera}.
This procedure enabled the collection of BP data across a wider range of values.

\section*{Appendix C}
For a subset of recordings (1699 samples) captured simultaneously by three cameras (Logitech C920n Webcam, Google Pixel 7, and Apple iPhone 14), we report the BP estimation performance of U-FaceBP.
TABLE~\ref{table:camera} shows the estimation performance of the model for each camera, using the same subject-independent 5-fold cross-validation as the main experiment.
The webcam achieves the highest estimation performance, whereas the smartphone cameras (Pixel 7 and iPhone 14) show slightly lower performance. 
This observation is consistent with previous findings~\cite{akamatsu2024calibrationphys}, where smartphone cameras exhibited lower heart rate estimation performance compared to the webcam.
This difference may be attributed to variations in device-specific image processing pipelines, color characteristics, and compression mechanisms, which can affect the quality of the extracted pulse waveforms.
Nevertheless, U-FaceBP maintains moderate correlation performance ($\geq$0.50) even on smartphone devices, indicating reasonable robustness across different camera types.

\setlength{\tabcolsep}{4pt}
\begin{table}[h]
\caption{BP estimation performance across three cameras.}
\vspace{-10pt}
\label{table:camera}
\begin{center}
\scalebox{1}{
        \begin{tabular}{lcccccc}  \noalign{\hrule height 1.3pt} \noalign{\smallskip}
        &  \multicolumn{5}{c}{FaceBP Dataset}  
        \\ \noalign{\smallskip}
        \cline{2-6} \noalign{\smallskip}
        & \multicolumn{2}{c}{SBP} & &  \multicolumn{2}{c}{DBP}
        \\ \noalign{\smallskip}
        \cline{2-3} \cline{5-6}
        \noalign{\smallskip}
        Camera &  MAE$\downarrow$ & Corr.$\uparrow$ & & MAE$\downarrow$ & Corr.$\uparrow$    \\ \noalign{\smallskip} \hline
        \noalign{\smallskip}
        Logitech C920n Webcam & 11.83 & 0.613 & & 8.57 & 0.579   \\
        Google Pixel 7 & 12.52 & 0.539 & & 8.94 & 0.517 \\
        Apple iPhone 14 & 12.89 & 0.500 & & 9.09 & 0.502 
        \\ \noalign{\smallskip} \noalign{\hrule height 1.3pt}
        \end{tabular}
        }
\end{center}
\end{table}
\setlength{\tabcolsep}{1.4pt}

\section*{Appendix D}
\label{app:cost}
Table~\ref{table:cost} shows the computational cost of our approach for training and inference.
The computational cost of training corresponds to 1-fold in the 5-fold cross-validation on the FaceBP dataset, and that of inference corresponds to 1 video.
The FLOPs results reflect multiple random sampling and sampling with MC dropout.
From the table, the training and inference times are short, and the inference can run in about two seconds even on a CPU environment.

\setlength{\tabcolsep}{4pt}
\begin{table}[h]
\caption{Computational cost (Training: 1-fold, Inference: 1 video).}
\vspace{-15pt}
\label{table:cost}
\begin{center}
\scalebox{1}{
\begin{tabular}{lcccc}
\noalign{\hrule height 1.3pt} \noalign{\smallskip}
 & rPPG & PPG & Image & Sum  \\ 
\noalign{\smallskip} \hline \noalign{\smallskip}
Training (eight A100 GPUs) & 119s & 450s & 229s & 798s   \\
Inference (one RTX 3060 GPU) & 0.127s & 0.271s & 0.0705s & 0.4685s \\
Inference (i7-10700 CPU) & 0.161s & 1.396s & 0.748s & 2.305s\\
\#Model Parameters & 0.444M & 1.702M & 23.516M & 25.662M \\
\#Sampling from face video & 60 & 10 & 3 & 73 \\
FLOPs & 13.2G & 143.5G & 86.2G & 242.9G  \\
\noalign{\smallskip} \noalign{\hrule height 1.3pt} \\
\end{tabular}
} \vspace{-10pt}
\end{center}
\end{table}
\setlength{\tabcolsep}{1.4pt}

\setlength{\tabcolsep}{4pt}
\begin{table*}[t]
\caption{Details of hyperparameter tuning.}
\vspace{-8pt}
\label{table:param}
\begin{center}
\scalebox{0.9}{
\begin{tabular}{lccc}
\noalign{\hrule height 1.3pt} \noalign{\smallskip}
Hyperparameter & Range & Final parameter & Selecting criterion   \\ 
\noalign{\smallskip} \hline \noalign{\smallskip}
$\alpha, \beta, \gamma$ & 1, 5, 10, 15 & $\alpha=5, \beta=10, \gamma=15$ & Tuned using validation data by balancing the scale of each loss term.  \\
\multirow{2}{*}{Learning rates} & \multirow{2}{*}{1e-3, 1e-4, 1e-5} & rPPG, PPG: 1e-3, Image: 1e-4 (FaceBP) & \multirow{2}{*}{Tuned using validation data by monitoring validation loss.}  \\
& & rPPG, Image: 1e-4, PPG: 1e-5 (MSPM, Fine-tuning) & \\
Batch size & 64, 128, 256 & 128 & Tuned using validation data by monitoring validation loss.  \\
$T$ & 5, 10, 30 & 10 & Determined from the perspective of computational cost. \\
\#Sampling from video & 3, 5, 10, 30 ,60 & rPPG: 60, PPG: 10, Image: 3 &  Determined from the perspective of computational cost. \\
\noalign{\smallskip} \noalign{\hrule height 1.3pt} \\
\end{tabular}
} \vspace{-10pt}
\end{center}
\end{table*}
\setlength{\tabcolsep}{1.4pt}

\section*{Appendix E}
\label{app:imp}
We explain the implementation details of the comparison methods and our U-FaceBP.
In order to perform a fair comparison between the methods, we make the implementation such as data preprocessing of the comparison methods and our U-FaceBP as similar as possible.

\subsection{Common Implementation}
\label{sec:imp-common}
As the ROI from face videos, we commonly use the area under the eyes (see Fig.~2~(b)).
The methods using PTT-related features, $i.e.$, S-Net, FS-Net, and U-FaceBP use multiple ROIs from the cheek, inner cheek, and forehead (see Fig.~2~(a)).
In the comparison methods, we extract rPPG signals using the approaches reported in each paper.
The rPPG signal is normalized to mean 0 and standard deviation 1, but a filtering process is not applied to the extracted rPPG signal (we confirm that rPPG signals without filtering generally perform better in BP estimation).
The BP values of the training data are normalized, and the estimated values in the test data are transformed back to the original scale using the mean and standard deviation of the training data.
In each epoch, samples of 5 seconds are randomly sampled for model training.
The sampling number is 60 for models using rPPG signals, 10 for models using PPG (or RGB) signals (U-FaceBP and Liu et al.~\cite{liu2024ensemble}), and 3 for models using face images (U-FaceBP).
The oversampling strategy of training data, number of epochs, optimization algorithm, batch size, and learning rate described in the experimental setup subsection are the same for all methods.
For the MSPM dataset (fine-tuning the models trained on the FaceBP dataset), learning rates are 1e-4, 1e-5, and 1e-4 for BP estimation from rPPG, PPG (or RGB) signals, and face images, respectively.
Note that the oversampling strategy is not implemented for BP estimation from face images (U-FaceBP), since the predicted BPs do not exhibit a tendency to concentrate around the mean BP values.
We also implement a scheduler that decreases the learning rate every 10 epochs, in which the rate of decrease is adjusted by monitoring the validation loss.

\begin{figure}[t]
\begin{center}
\includegraphics[scale=0.37]{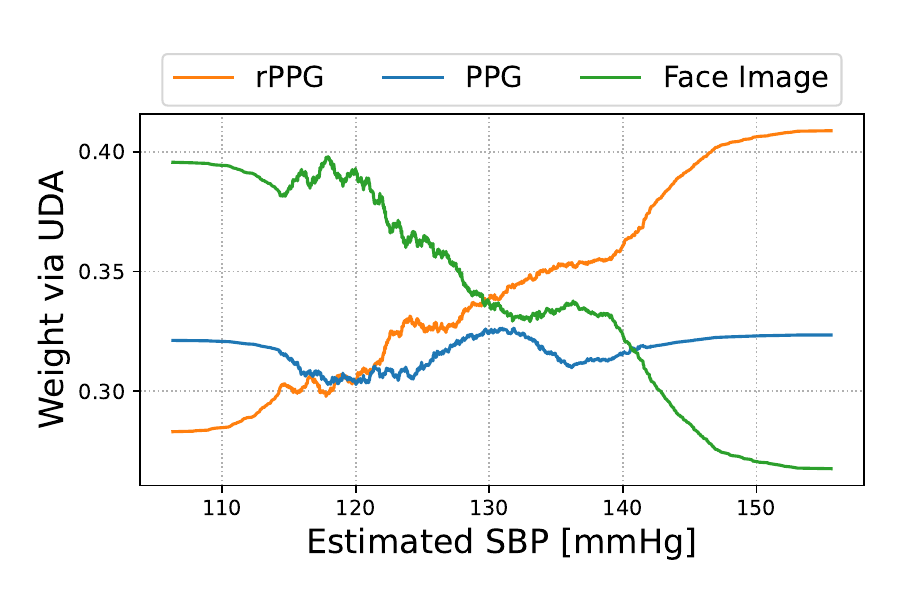}
\hspace{-5pt}
\includegraphics[scale=0.37]{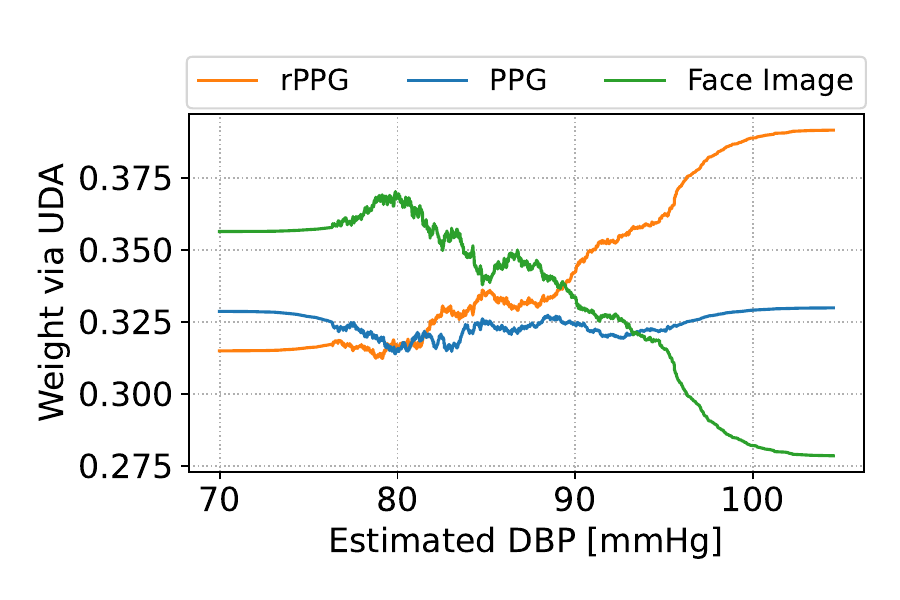}
\end{center}
\vspace{-15pt}
\caption{Weights of uncertainty-driven aggregator (UDA) for SBP and DBP on FaceBP dataset.}
\vspace{-10pt}
\label{fig:uda}
\end{figure}

\begin{figure*}[t]
\begin{center}
\includegraphics[scale=0.36]{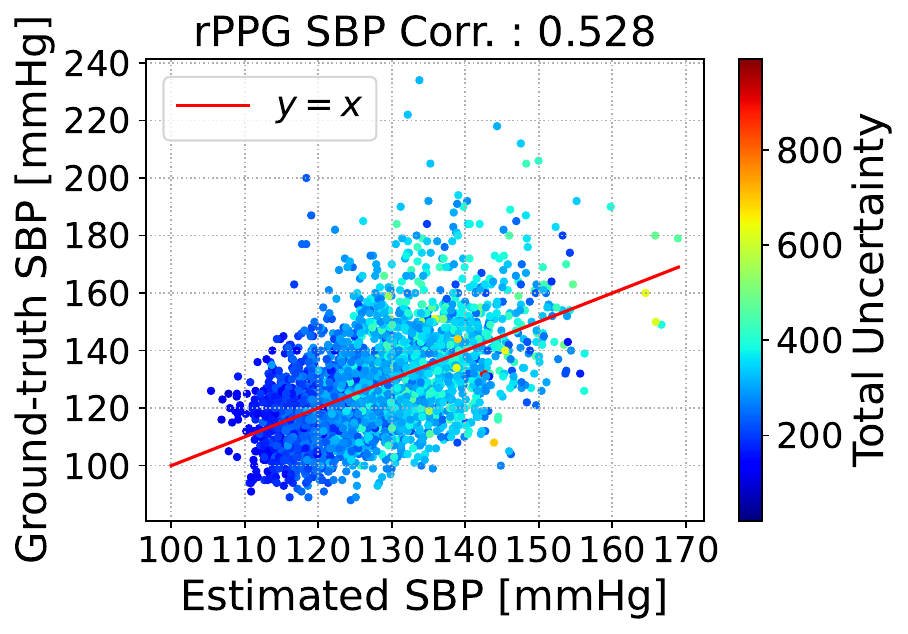}
\includegraphics[scale=0.36]{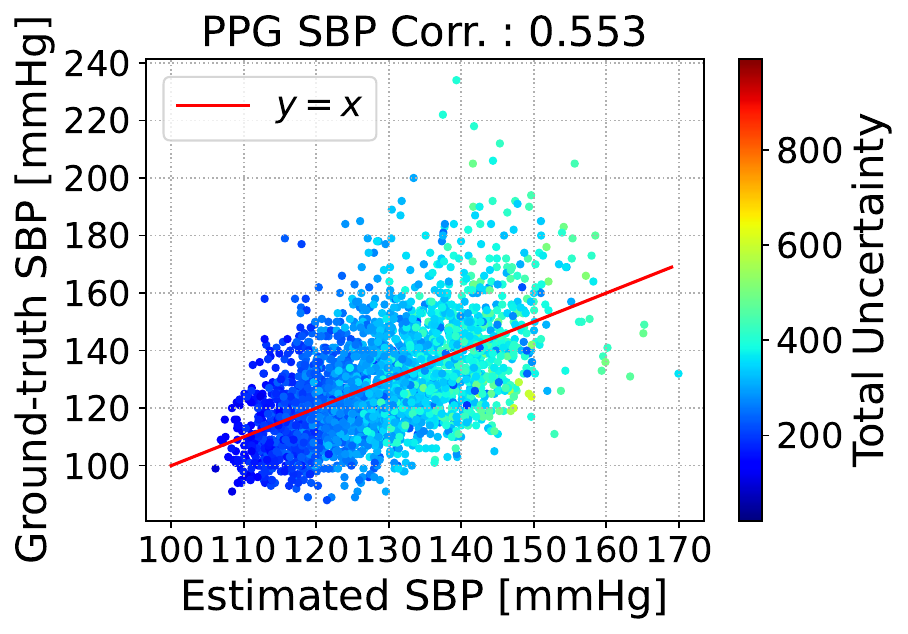}
\includegraphics[scale=0.36]{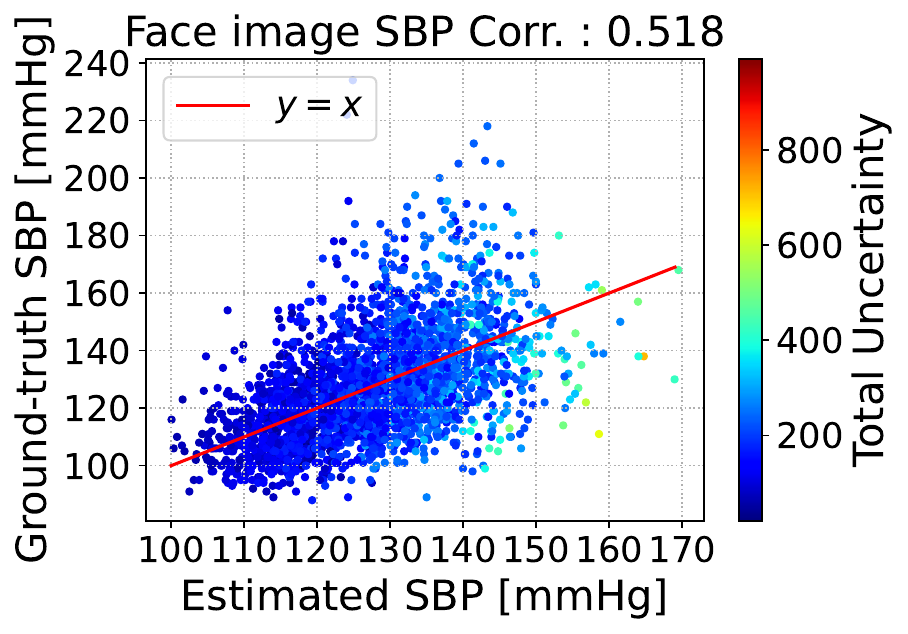}
\\
\includegraphics[scale=0.36]{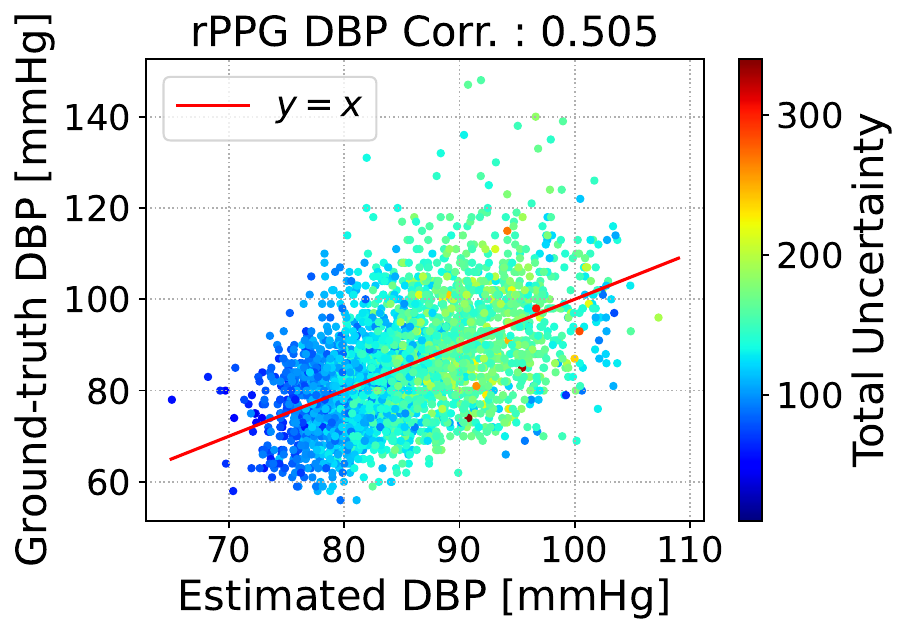}
\includegraphics[scale=0.36]{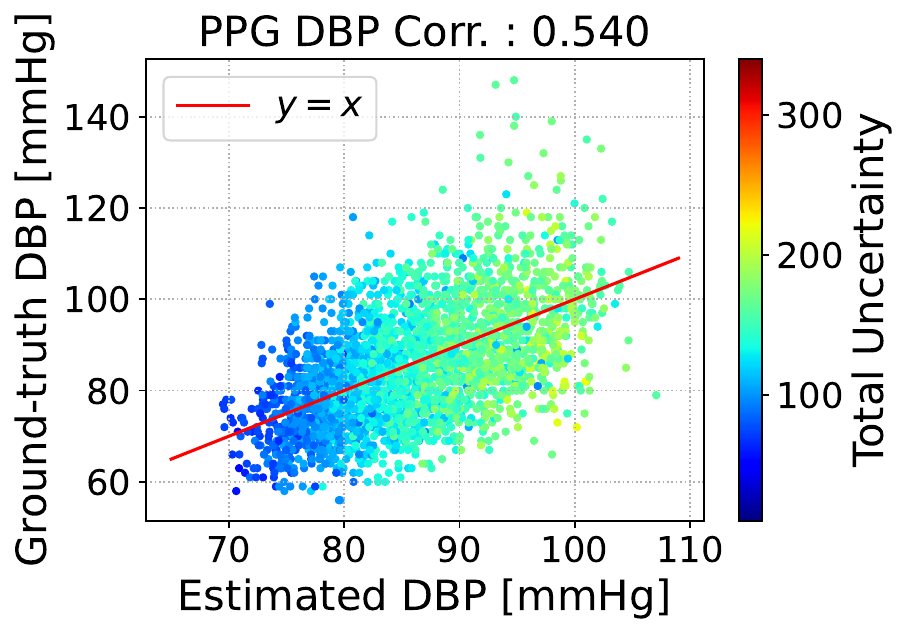}
\includegraphics[scale=0.36]{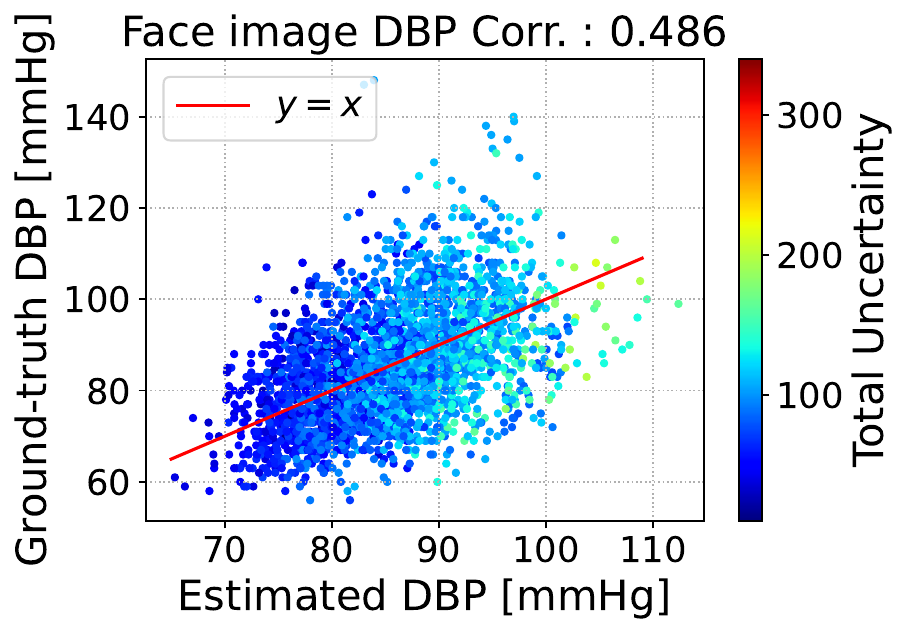}
\end{center}
\vspace{-10pt}
\caption{Correlation plots with total uncertainty for rPPG, PPG, and face image modalities.}
%\vspace{-5pt}
\label{fig:corr_uns}
\end{figure*}

\subsection{Comparison Methods}
\label{sec:imp-cm}
Mean Regressor is implemented to predict the mean SBP and DBP of the training data in each cross-validation.

Zhou et al.~\cite{zhou2019noninvasive} used body mass index (BMI) in addition to rPPG signals for BP estimation, but BMI was not acquired in the FaceBP dataset.
Therefore, we use the BMI estimated from the face image (this use is fair for comparison in terms of using only information obtained from the face for BP estimation).
Specifically, we use ResNet50~\cite{he2016deep} to predict BMI from face images as in a previous method~\cite{wu2022facial}, and train a model from ImageNet pre-training weights~\cite{deng2009imagenet} with the Illinois DOC labeled faces dataset~\footnote[6]{\url{https://www.kaggle.com/datasets/davidjfisher/illinois-doc-labeled-faces-dataset}}.
Then, the method by Zhou et al.~\cite{zhou2019noninvasive} is implemented to fit the model parameters using the least squares fitting algorithm according to the paper~\cite{zhou2019noninvasive}.

ResNet1D~\cite{schrumpf2021assessment} is implemented with the code in a previous study~\cite{gonzalez2023benchmark}.
AlexNet1D~\cite{schrumpf2021assessment} is implemented with reference to the original code~\cite{schrumpf2021assessment}.
To make a fair comparison with other methods, the model is trained from scratch (not transfer learning as in the above study~\cite{schrumpf2021assessment}).

S-Net~\cite{wu2022facial} is implemented according to the paper~\cite{wu2022facial}, where models are trained individually for SBP and DBP.
Compared to S-Net, FS-Net~\cite{wu2022facial} additionally uses heart rate (HR), heart rate variability (HRV), pulse transit time (PTT), and BMI (from face images).
HR and HRV are calculated using rPPG signals from the inner cheek, and PTT is calculated using rPPG signals from the cheek and forehead (see Fig.~2~(a)).
BMI is estimated in the same way as in the above implementation of the method by Zhou et al.~\cite{zhou2019noninvasive}.

CWT-CNN~\cite{trirongjitmoah2024assessing} is implemented according to the paper~\cite{trirongjitmoah2024assessing}.
Since the loss function for training is not provided, we use the mean squared error (MSE) loss, which is commonly used for BP estimation.

The method by Liu et al.~\cite{liu2024ensemble} is implemented according to the paper~\cite{liu2024ensemble}.
In terms of computational resources, the number of seeds $N$ in the ensemble method is set to 4, and each frame of face videos is resized to 36 $\times$ 36.

\subsection{U-FaceBP}
\label{sec:imp-pm}

Most of the implementation for U-FaceBP is explained in the experimental setup subsection, but here we describe the other details.
The original S-Net~\cite{wu2022facial} seems to train SBP and DBP individually, but our S-Net backbone (rPPG signals) predicts SBP and DBP simultaneously in a single model.
MC dropout is implemented in each residual block for S-Net (rPPG signals), each residual block except the first block for ResNet1D (PPG signals), and each residual block for ResNet50 (face images).
The dropout probability is 0.2 for S-Net, 0.5 for ResNet1D, and 0.5 for ResNet50.
For BP estimation from PPG signals, VPG signals are obtained by the first-order differences of PPG signals, and APG signals are obtained by the first-order differences of VPG signals, according to the implementation by Gonz{\'a}lez et al.~\cite{gonzalez2023benchmark}.
Table~\ref{table:param} summarizes the details on the hyperparameter tuning.

\section*{Appendix F}
\label{app:uda}
Figure~\ref{fig:uda} shows weights of uncertainty-driven aggregator (UDA) for SBP and DBP on the FaceBP dataset.
The weights are smoothed using a Savitzky–Golay filter to reduce local fluctuations.
In our UDA framework, we observe that the face image modality tends to receive higher weights for lower BP values. 
To further investigate this behavior, we present the correlation plots of each ensemble component in Fig.~\ref{fig:corr_uns}, annotated with total uncertainty. 
The plots indicate that the face image modality is capable of accurately estimating BP in the lower range ($e.g.$, SBP below 110 mmHg and DBP below 70 mmHg). 
In contrast, for higher BP values, the rPPG and PPG modalities—which are based on pulse wave information—receive greater weights than the face image modality.
This adaptive behavior aligns well with the observed performance of each modality across different BP ranges, indicating that UDA assigns weights in a reasonable and data-driven manner.
Overall, this demonstrates that UDA effectively balances appearance-based and physiological modalities depending on the target BP, supporting its role in enhancing overall estimation performance.

\bibliographystyle{IEEEbib}
\bibliography{refs}

\vfill

\end{document}